\documentclass{article}

\usepackage[nonatbib,preprint]{neurips_2021}
\usepackage[numbers]{natbib}

\usepackage[utf8]{inputenc} % allow utf-8 input
\usepackage[T1]{fontenc}    % use 8-bit T1 fonts
\usepackage{hyperref}       % hyperlinks
\usepackage{url}            % simple URL typesetting
\usepackage{booktabs}       % professional-quality tables
\usepackage{amsfonts}       % blackboard math symbols
\usepackage{nicefrac}       % compact symbols for 1/2, etc.
\usepackage{microtype}      % microtypography
\usepackage{xcolor}         % colors
\usepackage{algorithm}
\usepackage{algorithmic}
\usepackage{wrapfig} % NeurIPS .. for wrapping figures

\renewcommand{\P}{\ensuremath{\mathcal{P}}}

\usepackage{mathtools}
\usepackage{amsmath, amsthm, amssymb}
\DeclareMathOperator*{\argmin}{arg\,min}
\DeclareMathOperator{\Cov}{Cov}

\usepackage[inline]{enumitem}
\setlist*[enumerate,1]{%
  label=\arabic*),
}
\usepackage{cleveref}
\newtheorem{definition}{Definition}

\newtheorem{example}{Example}
\newtheorem{theorem}{Theorem}

\newcommand{\indep}{\perp \!\!\! \perp}
\newcommand{\EQ}{\ensuremath{\!=\!}}
\newcommand{\x}{\ensuremath{\!\times\!}}

\title{Debiasing classifiers: is reality at variance with expectation?}

\begin{document}

\author{%
  Ashrya Agrawal \\
  Birla Institute of Technology and Science\\
  Pilani, India\\
  \texttt{ashryaagr@gmail.com} \\
  \And
  Florian Pfisterer\\
  Ludwig-Maximilians-University \\
  M\"unich, Germany \\
  \texttt{florian.pfisterer@stat.uni-muenchen.de} \\
  \And
  Bernd Bischl \\
  Ludwig-Maximilians-University \\
  M\"unich, Germany \\
  \texttt{bernd.bischl@stat.uni-muenchen.de} \\
  \And
  Francois Buet-Golfouse \\
  J.P.\ Morgan \\
  London, United Kingdom\\
  \texttt{francois.buet-golfouse@jpmorgan.com} \\
  \And
  Srijan Sood \\
  J.P.\ Morgan AI Research \\
  New York, New York, USA \\
  \texttt{srijan.sood@jpmorgan.com} \\
  \And
  Jiahao Chen \\
  J.P.\ Morgan AI Research \\
  New York, New York, USA \\
  \texttt{jiahao.chen@jpmorgan.com} \\
  \And
  Sameena Shah \\
  J.P.\ Morgan AI Research \\
  New York, New York, USA \\
  \texttt{sameena.shah@jpmorgan.com} \\
  \And
  Sebastian Vollmer \\
  University of Warwick \\
  Warwick, United Kingdom\\
  \texttt{svollmer@warwick.ac.uk} \\
}

\maketitle

% Points to make:
% - Theoretical and Empirical Study
% -> Variance increases through debiasing
% -> Extra variance makes model selection difficult
% - Debiasing optimizes a different optimization problem than decreasing fairness
% -> Partial debiasing is sensible

% Many methods for debiasing classifiers have been proposed, but their effectiveness in practice remains unclear.
% In practice, we observe that debiasing can improve or worsen out-of-sample accuracy and fairness,
% while often resulting in significantly increased variance of performance estimates -- in short, debiasing often generalizes poorly to out-of-sample data.
% We study the generalizability of those methods theoretically, as well as in an empirical study.
% After framing this problem as a trade-off between predictive performance and fairness
% we theoretically study the corresponding bias-variance decomposition
% and find that the variance in estimation arises from class imbalances with respect to both the outcome and the protected classes.
% We corroborate those findings in an empirical study, focussing on practical consequences.
% Specifically, we propose and analyse partial debiasing as a tool to arrive at better trade-offs and provide further guidance for practical use of debiasing techniques.

\begin{abstract}
We present an empirical study of debiasing methods for classifiers,
showing that debiasers often fail in practice to generalize out-of-sample,
and can in fact make fairness worse rather than better.
A rigorous evaluation of the debiasing treatment effect requires
extensive cross-validation beyond what is usually done.
We demonstrate that this phenomenon can be explained as
a consequence of bias-variance trade-off, with an increase in variance
necessitated by imposing a fairness constraint.
Follow-up experiments validate the theoretical prediction
that the estimation variance depends strongly on the base
rates of the protected class.
Considering fairness--performance trade-offs justifies the
counterintuitive notion that partial debiasing
can actually yield better results in practice on out-of-sample
data.
\end{abstract}

\section{Introduction}

Artificial intelligence and machine learning (AI/ML) are now used
for many high-stakes decision-making processes at scale \citep{Mehrabi2019,Rudin2019},
such as credit decisions \citep{Chen2018,Turner2019},
medical diagnoses \citep{Topol2019},
and criminal sentencing \citep{compas,corbettcompas,richardcompas}.
In these use cases, unfairness is not just an ethical concern,
but has legal and regulatory dimensions as well \citep{Barocas2016,Chen2018,Xiang2019,Kurshan2020,ico}.
As such, regulators have signalled their interests in detecting and remediating bias
in these real-world applications \citep{ico,usrfi}.

% How do we measure and mitigate bias?

Bias can originate from any part of the machine learning modeling process,
ranging from exclusionary biases \citep{Barocas2016,Buolamwini2018,Richardson2019} in training data,
to problem definitions or feedback cycles that reinforce historical and systemic discrimination
\citep{Freeman2017,liu19delayed,barabas20,decolonialai}.
% The standard workflow for mitigating biases in a machine learning model comprises two parts: first,
%  the identification of relevant fairness metrics \citep{narayanan2018translation,Verma2018}
% and second, the selection of an appropriate method by which to debias the model.
To address bias in a model, one must first identify the relevant fairness metrics \citep{narayanan2018translation,Verma2018}
and then select a method to debias the model with respect to those metrics.
However, both aspects are challenging in practice.
It is not always obvious which fairness definitions are relevant for a particular application \citep{compas,richardcompas},
and remediating bias usually comes at a cost.
% despite the recommendations that are based on the resulting interventions \citep{Rodolfa2020-kn},
For example, a credit decisioning model has to be accurate in order to be profitable,
which motivates fairness notions like equality of opportunity.
At the same time, there are reputational and regulatory risks associated with bias in incorrect decisions,
leading to considerations of equalized false negative rate and equalized false positive rate \citep{Hardt2016-zi}.
These different definitions of fairness cannot be satisfied simultaneously due to well-known impossibility theorems \citep{kleinberg2016inherent,chouldechova2017fair}.
Furthermore, a debiased model will not be used in practice if its performance degrades too much.
Therefore, in practice, we have to consider not only fairness--fairness trade-offs,
but also the fairness--performance trade-offs to determine the best debiased model \citep{Pleiss2017,fact}.

% Context for our paper

\paragraph{Assumptions.}\label{para:assumptions}
The setting for our paper assumes that
\begin{enumerate*}
\item membership in protected classes is fully known, ignoring practical concerns to the contrary \citep{Chen2019,Kallus2020},
\item all relevant fairness and performance metrics can be clearly identified at the outset, and
\item remediation is only at single point in time, ignoring time-evolving concerns \citep{liu19delayed}.
\end{enumerate*}
% In this paper, we assume that
% and all relevant sociotechnical concerns have been identified from domain knowledge,
% resulting in the need for simultaneous consideration of multiple fairness metrics and performance metrics - a simplified abstraction of the aforementioned real world use cases.
% And furthermore, we focus on
Despite this restricted setting, we find that the practicalities of debiasing are already sufficiently rich for in-depth study.
% but did not explain why.
% although no explanation was provide.
% ,
% focusing primarily on the metrics of disparate impact and demographic parity \citep{Calders2010}.
Other work identified technical challenges resulting from the lack of native support
for fairness or debiasing concerns in major machine learning software libraries,
but do not consider variance or sensitivity issues \citep{Biswas2020-jk}.
More recently, \citet{rodolfa2020machine} show that trade-offs between recall parity and precision$@k$ are often small in
real-world projects.

Given this body of work, we were therefore surprised to it is surprising to see results like those in \Cref{fig:matrix_scatter},
which suggest that classifiers can exhibit any combination of improved or worsened fairness,
and also improved or worsened performance, after using standard debiasing algorithms.
In this paper, we show that out-of-sample generalization error is responsible for the fluctuations observed in the aforementioned figure,
and that careful estimation of such error is essential for proper evaluation of debiasing methods.
While previous works have studied distributionally robust optimization for fairness \citep{2020LinBeyondTraining} and data-dependent constraint generalization \citep{Cotter2019WellGeneralizing}, we focus on the generalization of fairness algorithms.

\paragraph{Our Contributions.}
% \begin{enumerate}
% % \begin{enumerate*}[series=contribs]
% \item
In \Cref{sec:generaldebiasers}, we show how to generalize existing debiasers to apply to fairness metrics other than
what they were originally defined for.
We introduce generalized reweighing which can apply to other fairness definitions beyond demographic parity
and identify fairness definitions for which reweighing cannot generalize to.
We also introduce a new NLinProg debiaser which generalizes the equalized odds debiaser,
and is capable of handling multiple fairness and performance metrics simultaneously.
% \item
In \Cref{sec:benchmark}, we present a detailed empirical study across nine different models, showing that debiasing methods generally fail to achieve perfect fairness in out-of-sample measurements, and produce large variance in the actual metrics,
and tend to overfit on training data.
% \end{enumerate*}
% \begin{enumerate}[resume=contribs,leftmargin=*]
% \item
In \Cref{sec:learningtheory}, we present our main theoretical result, \Cref{prop:var}, showing that the
fluctuations we observed empirically can be attributed to bias-variance trade-off.
% \item
In \Cref{sec:imbalance}, we verify a prediction from this analysis, that the ability to debias varies with the base rate of
the protected class.
% \item
In \Cref{sec:partial_debiasing}, we show how an explicit consideration of the fairness--performance trade-offs motivates
the notion of partial debiasing.
We also show experimentally the somewhat counter-intuitive result that a \textit{partial}
debiasing treatment can actually yield classifiers with better out-of-sample fairness.
We introduce other related work throughout the exposition of this paper, in lieu of a dedicated section.

\paragraph{Notation.}
\textbf{Sets.} In general, calligraphic letters like $\mathcal A$ denote a set,
capital letters $A$ denote a variable that is an element of a set $\mathcal A$,
and small letters $a$ denote a value that the variable $A$ can take.
Let
$S\in\mathcal S = \{0,1\}$ be a binary \textit{protected class},
$X\in\mathcal X$ be some set of \textit{features} that explicitly excludes $\mathcal S$,
    % so that $\mathcal X \cap \mathcal S = \varnothing$,
$Y\in\mathcal Y = \{0,1\}$ be a binary \textit{outcome variable},
and $\hat Y\in\mathcal Y$ be an \textit{estimator} for $Y$.
While we specialize to the cases of binary $\mathcal Y$ and $\mathcal S$ for the ease of presentation,
our results generalize to larger finite classes.
Furthermore, let
$Z = (X, Y) \in \mathcal Z = \mathcal X\x{}\mathcal Y$ and
$W = (X, Y, S) \in \mathcal W = \mathcal X\x{}\mathcal Y\x{}\mathcal S$,
% , and
% $\hat {\mathcal Y}_f \subseteq \mathcal Y $ be the range of prediction for some classifier $f$,
% with the subscript $f$ dropped when obvious from context.
% , and similarly for $X$ and $Y$.
Additionally, define
$\mathcal D \in \mathcal W^n$ to be \textit{in-sample (training) data} with $n$ points,
$\mathcal D^\star \in \mathcal W^{n^\star}$ to be out-of-sample \textit{(testing) data} with $n^\star$ points,
and $\Delta^k = \{z\in\mathbb R^{k+1}: z\ge 0, \Vert z \Vert_1 = 1 \}$ be the standard non-negative simplex of dimensionality $k$.
\textbf{Classification functions.} Let
$\mathcal F : \mathcal X\x{}\mathcal S \rightarrow \mathcal Y$
be the function space of $\mathcal S$-\textit{aware classifiers},
where each element $f \in \mathcal F$ is a classification function,
and $\mathcal F_0 : \mathcal X \rightarrow \mathcal Y$
be the function space of $\mathcal S$-\textit{oblivious classifiers}.
Each $\mathcal S$-oblivious classifier $f_0 \in F_0$ has a 1:1 relation
to a trivial $\mathcal S$-aware classifier
$f\in F: f(x, s) = f_0(s)$ which simply ignores the $s$ argument.
We differentiate between aware and oblivious classifiers only where necessary.
Also, let
$\mathcal H \subseteq \mathcal F$ be some family of classifiers, and
$\textrm{id}_\mathcal{A}: \mathcal A \rightarrow \mathcal A$ be the identity function over the set $\mathcal A$.
\textbf{Debiasing functions.} Let
$g_\textrm{pre}: \mathcal X\x{}\mathcal S \rightarrow \mathcal X$ be a pre-processing debiasing function,
$g_\textrm{post}: \mathcal Y\x{}\mathcal S \rightarrow \mathcal Y$ be a post-processing debiasing function,
and
$G: \mathcal F \rightarrow \mathcal F$ be an in-processing debiaser, which is a higher-order function
that is an endomorphism over $\mathcal F$.
\textbf{Loss functions.} Let
$\ell : \mathcal F \x{} \mathcal W \rightarrow \mathbb R^+_0$
be a performance loss such as the hinge or binomial deviance,
and $\phi_h : \mathbb R^2 \rightarrow = R^+_0$ be a loss function associated with the fairness definition $h$.
\textbf{Metrics.} Let
$\gamma : \mathcal F \x{} \mathcal W^{\mathbb N} \rightarrow [0, 1]$, $\gamma(f, \mathcal D)$ be the accuracy of the classifier $f$ on the data set $\mathcal D$,
and $\tau_{h} : \mathcal F \x{} \mathcal W^{\mathbb N} \rightarrow [0, 1]$ , $\tau_h(f, \mathcal D)$ the fairness metric as defined in \Cref{def:ratiometric}
corresponding to the fairness definition $h$.
When clear from context, the arguments $f$ and $\mathcal D$ will be dropped for brevity.

\subsection{Fairness definitions \& metrics}
\label{sec:fairness_definitions}

\begin{table*}[ht]
\centering
% \scalebox{0.8}{
\begin{tabular}{|l|c|}
\hline
Fairness metric & Equality statement\tabularnewline
\hline
\hline
Equalized false omission rate (EFOR) \citep{richardcompas} & $\Pr(Y=1|\hat{Y}=0,S=s)=\Pr(Y=1|\hat{Y}=0)$\tabularnewline
\hline
Predictive parity (PP) \citep{chouldechova2017fair}
& $\Pr(Y=1|\hat{Y}=1,S=s)=\Pr(Y=1|\hat{Y}=1)$\tabularnewline
\hline
Demographic parity (DP) \citep{Calders2010}
& $\Pr(\hat{Y}=1|S=s)=\Pr(\hat{Y}=1)$\tabularnewline
\hline
  Equalized false negative rate (EFNR) \citep{chouldechova2017fair}
& $\Pr(\hat{Y}=0|Y=1,S=s)=\Pr(\hat{Y}=0|Y=1)$\tabularnewline
\hline
Predictive equality (PE) \citep{chouldechova2017fair}
& $\Pr(\hat{Y}=1|Y=0,S=s)=\Pr(\hat{Y}=1|Y=0)$\tabularnewline
\hline
Equality of opportunity (EOp) \citep{Hardt2016-zi}
& $\Pr(\hat{Y}=1|Y=1,S=s)=\Pr(\hat{Y}=1|Y=1)$\tabularnewline
\hline
Equalized odds (EOd) \citep{Hardt2016-zi} & EOp and PE\tabularnewline
\hline
\end{tabular}
% }
\caption{Group fairness definitions used in this paper.}
\label{tab:fairness}
\end{table*}

Many technical definitions of fairness exist and they have been reviewed elsewhere \citep{narayanan2018translation,Verma2018,richardcompas,fact}.
We present only the definitions of fairness that we will study in this paper in \Cref{tab:fairness}.
In addition to choosing a suitable fairness definition,
we also have to choose some loss function, $\phi$,
to quantify the discrepancy from perfect fairness.
One such function is the Calders-Verwer gap \citep{Calders2010}
$\Delta_\textsc{DP} = \Pr(\hat Y\EQ1 | S\EQ1) - \Pr(\hat Y\EQ1 | S\EQ0)$,
which is simply the difference of the two sides of the equation that define demographic parity,
and vanishes when perfect fairness exists.
In addition to absolute differences, other metrics based on ratios, relative differences,
or other more complicated losses have have been proposed.
In this paper, we focus on symmetrized ratio metrics
as defined in \Cref{def:ratiometric}.
% of the form
% \begin{equation}
% \tau_{{}_\textsc{DP}} = \min\left(
% \frac {\Pr(\hat Y\EQ1 | S\EQ1)} {\Pr(\hat Y\EQ1 | S\EQ0)},
% \frac {\Pr(\hat Y\EQ1 | S\EQ0)} {\Pr(\hat Y\EQ1 | S\EQ1)}
% \right),
% \label{eq:tau}
% \end{equation}
% %
% and similarly for other group fairness measures.
% By definition, these $\tau$s
% % are constrained to
% lie within the unit interval $0 \le \tau \le 1$.
% The symmetry with respect to interchanging $S=0$ and $S=1$
% removes the need to assume that there exists a protected class $S=0$ that is generally privileged.
% Considering these $\tau$s
% allows us to compare different metrics across different datasets.

\subsection{Pre-, in- and post-processing methods for debiasing}
\label{sec:debiasing_methods}

\begin{wrapfigure}{r}{0.5\textwidth}
\centering
\includegraphics[width=0.4\textwidth]{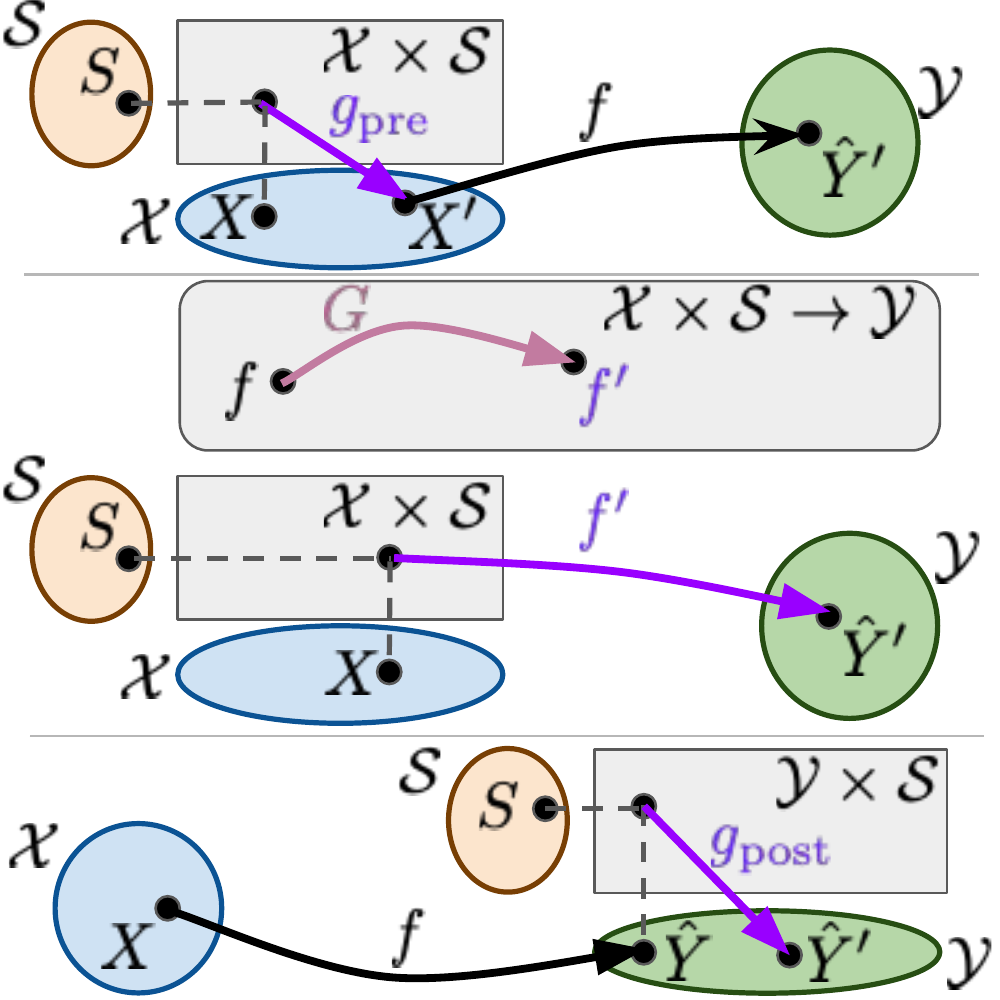}
\caption{Overview of debiasing methods as described in \Cref{sec:debiasing_methods}.
Top: pre-processing methods.
Middle: in-processing methods.
Bottom: post-processing methods.
}
\label{fig:intro_debiasing}
\end{wrapfigure}

Toolkits such as Aequitas \citep{aequitas}, IBM AI Fairness 360 \citep{aif360},
Microsoft Fairlearn \citep{fairlearn}, and Amazon SageMaker Clarify \citep{clarify}
provide many debiasing algorithms.
These algorithms are traditionally classified as pre-processing, in-processing and post-processing methods,
which are depicted at the functional level in \Cref{fig:intro_debiasing}.
In this section, all primed quantities have been debiased.
A \textit{pre-processing debiaser} first transforms the input features $\mathcal X$ using some function $g_\textrm{pre} : \mathcal X \x{} \mathcal S \rightarrow \mathcal X$,
then feeds the transformed features as input to an oblivious classifier $f: \mathcal X \rightarrow \mathcal Y$.
The debiased classifier is then the composition $f^\prime : \mathcal X \x{} \mathcal S \rightarrow \mathcal Y$, $f^\prime = f \circ g_\textrm{pre}$.
A \textit{post-processing debiaser} takes the output of some oblivious classifier, $\hat Y \in \mathcal Y$,
then transforms this output using some function $g_\textrm{post} : \mathcal Y \x{} \mathcal S \rightarrow \mathcal Y$,
The debiased classifier is then the composition
$f^\prime : \mathcal X \x{} \mathcal S \rightarrow \mathcal Y$,
$f^\prime = g_\textrm{post} \circ (f \x{} \textrm{id}_\mathcal{S})$,
where $\textrm{id}_\mathcal{S}$ is the identity function over protected class.
% (the projection of $\mathcal{X} \x{} \mathcal{S} $ onto $ \mathcal{S}$ ),
% defined element wise by $\hat Y = f(X)$, $f^\prime(X, S) = g_\textrm{post}(\hat Y, S)$.
%
Finally, an \textit{in-processing debiaser} transforms some oblivious classifier $f$ into an $\mathcal S$-aware but debiased classifier $f^\prime = G(f)$, using the function-to-function mapping $G:(\mathcal X \rightarrow \mathcal Y) \x{} \mathcal W \rightarrow (\mathcal X \x{} \mathcal S \rightarrow \mathcal Y)$.
% The resulting debiased classifier is then the composition
% $f^\prime : \mathcal X \x{} \mathcal S \rightarrow \mathcal Y$,
% $f^\prime = G(f)$,
In general, the resulting classifier cannot be written as a function composition involving the original oblivious classifier $f$.
Some in-processing debiasers like prejudice removal \citep{prejudiceremover} further require that
the debiased classifier $f^\prime$ be $\mathcal S$-oblivious, which is equivalent to the defining
the debiased classifier $f^\prime(X, S) = f(X)$ to be independent of $S$ always.
In other words, pre-processing debiasers transform the features $\mathcal X$,
post-processing debiasers transform the predictions $\hat {\mathcal Y}$,
and
in-processing debiasers transform the classifiers $f$.

We conclude the introduction with two debiasing algorithms
as illustrations of the general principle.
\textbf{Reweighing} (RW) is a pre-processing debiaser
introduced to enforce demographic parity (DP) \citep{kamiran2012data}.
Since DP is satisfied when $Y$ and $S$ are independent,
% since
% $
% \Pr(\hat Y\!=\!y | S\!=\!s) = {\Pr(\hat Y\!=\!y, S\!=\!s)}/{\Pr(S\!=\!s)} = \Pr(\hat Y\!=\!y).
% $
% where the first equality is by definition of the conditional probability,
% and the second equality follows from the statistical independence assumption.
reweighing assigns each data point $i$ a weight
% $w_{\textrm{DP},i} = $
% \vspace{-6pt} % Commented this for NeurIPS
% \begin{equation}
$
w_{\textrm{DP},i} = {\Pr(\hat Y\!=\!y_i) \Pr(S\!=\!s_i)} / {\Pr(\hat Y\!=\!y_i, S\!=\!s_i)}
$, %= \frac{\Pr(\hat Y\!=\!y_i)} {\Pr(\hat Y\!=\!y_i | S\!=\!s_i)},
\label{sec:reweighing}
% \vspace{-6pt} % Commented this for NeurIPS
% \end{equation}
altering the measure associated with the sampled distribution of $(Y, S)$
to match what would be expected from statistical independence.
% The resampling method simply resamples a new data distribution according to these weights $w$,
% to be used in classifiers that do not support arbitrary weights on data points.
% \subsection{Equalized odds post-processing}%
\label{sec:eqodds}
\textbf{Equalized odds} (EOd) is a post-processing debiaser \citep{Hardt2016-zi,Pleiss2017}
that calculates probabilities
$\Pr(\hat Y^\prime | \hat Y, S)$
% $\Pr(\hat Y^\prime | \hat Y = y, S = s)$ for each $y \in \mathcal Y$ and $s \in \mathcal S$
that the predictions $\hat Y$ should be flipped to yield the debiased predictions $\hat Y^\prime$ that satisfy equalized odds fairness, while having $\hat Y'$ as close as possible to $\hat Y$.
% The
% % consists of solving the following linear program
% % involving equalized odds fairness (EOd) between the original predictions $\hat Y$
% % and the debiased predictions $\hat Y^\prime$: %as a constraint:
% % % ,
% % % but making the substitution of variables $(Y, \hat Y) \gets (\hat Y, \hat Y^\prime)$ :
% \begin{equation}
% \label{eq:eqodds_debiaser}
% \begin{split}
% \min_{\hat Y^\prime}\; & \mathbb E \ell (\hat Y^\prime, \hat Y) \\
% \textrm{s.t. } & \Pr(\hat Y^\prime\!=\!1 | S\!=\!0, \hat Y\!=\!y) = \Pr(\hat Y^\prime\!=\!1 | S\!=\!0, \hat Y\!=\!y) \;\\ & \forall y\in\mathcal Y, \\
%               & \sum_{s\in\mathcal S} \Pr(\hat Y^\prime\!=\!1 | S\!=\!s, \hat Y\!=\!y) = 1 \;\forall y\in\mathcal Y, \\
% \textrm{and }  & \Pr(\hat Y^\prime\!=\!1 | S\!=\!s, \hat Y\!=\!y) \ge 0 \;\forall y\in\mathcal Y, s\in\mathcal S,
% \end{split}
% \end{equation}
% %
% where the last two conditions enforce that $\hat Y^\prime$ follows a valid probability distribution,
% and $\ell: \mathcal Y^2 \rightarrow \mathbb R$ is an unspecified loss function in \citep{Hardt2016-zi},
% but is assumed to be effectively $-\gamma$ in the code of \citep{Pleiss2017},
% where $\hat Y$ takes the place of the ground truth $Y$ in the definition of accuracy.
% %%%% End commenting for NeurIPS version

\section{Generalized debiasers}
\label{sec:generaldebiasers}

The reweighing pre-processor and equalized odds post-processsor
are specialized to specific fairness definitions,
demographic parity and equalized odds respectively.
In this section, we show how these debiasers can be
generalized to other fairness definitions.

\subsection{Generalized reweighing for pre-processing}

The reweighting pre-processor of \Cref{sec:reweighing} can be easily extended to some, but not all,
other definitions of group fairness.
For example, considering $\hat Y \indep S | Y\EQ1$ instead of $\hat Y \indep S$
gives an immediate generalization of reweighing for equality of opportunity (EOp) instead of DP.
As EOp fairness requires
% \vspace{-6pt} % Commented this for NeurIPS
% \begin{equation*}
$
\Pr(\hat Y\EQ1 | S = 0, Y = 1) = \Pr(\hat Y\EQ1 | S = 0, Y = 1),
$
% \end{equation*}
the corresponding reweighing scheme is simply
% \vspace{-6pt} % Commented this for NeurIPS
% \begin{equation}
$w_{\textrm{EOp}, i} = {\Pr(\hat Y = y_i)} / {\Pr(\hat Y = y_i | S = s_i, Y\EQ1)}.
$
% \label{eq:reweighing_eop}
% \vspace{-6pt} % Commented this for NeurIPS
% \end{equation}
%
However, there is no such reweighing scheme for equalized odds (EOd), which requires that both EOp and PE hold.
Each equation demands its own reweighing scheme, with the first as before and the second as in
% \vspace{-6pt} % Commented this for NeurIPS
% \begin{equation}
$
w_{\textrm{PE}, i} = {\Pr(\hat Y = y_i)} / {\Pr(\hat Y = y_i | S = s_i, Y\EQ0)},
$
% \label{eq:reweighing_neop}
% \vspace{-6pt} % Commented this for NeurIPS
% \end{equation}
%
which will in general differ from the weights $w_{\textrm{EOp}, i}$.
% defined in \cref{eq:reweighing_eop,eq:reweighing}.
Thus, reweighing as a method for exact debiasing works for
neither composite fairness definitions that require multiple equality constraints,
nor situations requiring multiple fairnesses to be satisfied simultaneously.
It is therefore natural to consider the possibility of some interpolation scheme
between different weighting schemes. We will revisit this idea later in \Cref{sec:partial_debiasing}.

\subsection{Nonlinear programs for post-processing (NLinProg; NLP)}%
\label{sec:NLinProg}

We now introduce \textbf{NLinProg} (NLP),
a generalization of the equalized odds post-processor
to allow for arbitrary combinations of group fairnesses
to be debiased simultaneously.

\begin{algorithm}
\caption{The NLinProg post-processing debiaser}
\label{alg:nlinprog}
\renewcommand{\algorithmicrequire}{\textbf{Input: }}
\renewcommand{\algorithmicensure}{\textbf{Output: }}
\algorithmicrequire{Predictions $\hat Y$, protected class $S$, performance losses $\{\ell^{(i)}\}_i$, and fairness losses $\{\phi^{(i)}\}_i$}. \\
\algorithmicensure {Debiased predictions $\hat Y^\prime$}. \\
\begin{algorithmic}[1]
% \DontPrintSemicolon
\STATE Compute the solution $z = (\Pr(\hat Y^\prime\!=\!y^\prime, \hat Y\!=\!y | S\!=\!s))_{y^\prime, y, s}$ to the PFOP \eqref{eq:pfop}. \\
\STATE For each prediction $\hat Y = y$ with protected class label $S = s$,
choose a corresponding debiased prediction $\hat Y^\prime = y^\prime$ with probability
$\Pr(\hat Y^\prime\!=\!y^\prime | \hat Y\!=\!y, S\!=\!s)$.
\end{algorithmic}
\end{algorithm}

\begin{definition}
The \textbf{performance--fairness optimality problem} (PFOP) is to determine the
fairness-confusion tensor (FACT) $z = (TP_1, FN_1, FP_1, TN_1, TP_0, FN_0, FP_0, TN_0)/N$  \citep{fact} that solves:
\begin{equation}
\argmin_{z\in \Delta^7}
\sum_i \mu_i \ell^{(i)} (z)
+
\sum_j \lambda_j \phi^{(j)} (z),
\label{eq:pfop}
\end{equation}
% where
% is the unraveled fairness-confusion tensor (FaCT)
% formed from the confusion matrices for the $S=1$ and $S=0$ classes stacked together,
where $TP_0/N = \Pr(\hat Y^\prime\EQ1, \hat Y\EQ1, S\EQ0)$ is the normalized true positive entry
for $S\EQ0$,
and similarly for the other entries of $z$,
$\Delta^7 = \{z\in\mathbb R^8: z\ge 0, \Vert z \Vert_1 = 1 \}$ is the standard non-negative simplex,
$\ell^{(i)} : \Delta^7 \rightarrow \mathbb R^+_0$ is some performance loss with corresponding Lagrange multiplier $\mu_i$, and
$\phi^{(i)} : \Delta^7 \rightarrow \mathbb R^+_0$ is some fairness loss with corresponding Lagrange multiplier $\lambda_j$.
% As above for the EOd post-processor, we define the entries of $z$ with respect to $(\hat Y^\prime,\hat Y)$ rather than the usual $(\hat Y | Y)$, so that
\end{definition}
We implement \Cref{alg:nlinprog} in the JuMP \citep{jump} framework for the Julia programming language \citep{julia},
which uses Ipopt \citep{ipopt} for interior point optimization.
The MIT-licensed open source implementation is available on GitHub.%
\footnote{URL redacted for double-blind peer review.}%

Unless otherwise specified,
our subsequent experiments specialize to
one accuracy loss $\ell^{(1)}(z) = 1 - \gamma(z)$,
where $\gamma(z) = \sum_i (TP_i + TN_i) / N$ is the usual definition of accuracy,
and fairness loss $\phi^{(1)}(z) = 1 - \tau_{h}(z)$, where $\tau_h$ is a quantity
we will now define.

\begin{definition}\label{def:ratiometric}
For a FACT $z$, define
$z_{|S\EQ{}s} = (TP_s, FN_s, FP_s, TN_s)/N$ as the restriction of $z$ to entries corresponding to $S\EQ{}s$.
Let $h: [0,1]^4 \rightarrow \mathbb R$ be a group fairness expressible as a constraint $h(z_{|S\EQ{}1}) = h(z_{|S\EQ{}0})$.
Then, the \textbf{symmetrized fairness gap} for the fairness $h$ at $z$ is $\Delta_h(z) = |h(z_{|S\EQ{}1}) - h(z_{|S\EQ{}0})|$, and the
\textbf{symmetrized ratio metric} for $h$ evaluated at $z$ is
% \begin{equation}
$
\tau_{h}(z_{|S\EQ{}0}, z_{|S\EQ{}1}) = \min\left({h(z_{|S\EQ1})}/{h(z_{|S\EQ0})}, {h(z_{|S\EQ0})}/{h(z_{|S\EQ1})} \right).
$
% \label{eq:tau}
% \end{equation}
\end{definition}
It is easy to show that $\tau_{h} \in [0, 1]$; we omit the proof of this simple fact.
Furthermore, $\tau_{h}$ is symmetric in its arguments, which
removes the need to assume that either class is generally privileged.
Where clear from context, we will (with abuse of notation) also write the above as
$\tau_{h}(z)$.

\begin{example}
Demographic parity $\Pr(\hat Y\EQ1 | S\EQ1) = \Pr(\hat Y\EQ1 | S\EQ0)$
can be expressed as $h_\textsc{DP}(z_{|S\EQ{}1}) = h_\textsc{DP}(z_{|S\EQ{}0})$ with the function
% \begin{equation}
$
h_\textsc{DP}(z_{|S\EQ{}s}) = \Pr(\hat Y\EQ{}s | S\EQ{}s)
= {(TP_s + FP_s)}/{(TP_s + FP_s + FN_s + TN_s)}.
$
% \label{eq:h-dp}
% \end{equation}
\end{example}
%
% and similarly for other group fairness measures.
% By definition, these $\tau$s
% are constrained to
% lie within the unit interval $0 \le \tau \le 1$.
% The symmetry of the corresponding $\tau_\textsc{DP}$ with respect to interchanging $S\EQ0$ and $S\EQ1$
% removes the need to assume that there exists a protected class $S\EQ1$
% that is generally privileged.

We do not recommend NLinProg for general use---as we will see in \Cref{sec:benchmark},
its performance is generally Pareto suboptimal, in that it yields neither the most fair
classifiers nor the most accurate classifiers. However, for our experiments,
NLinProg serves as a useful construct for investigating the general behavior of post-processing methods.

\section{Empirical evaluation of debiasers}
\label{sec:benchmark}

\paragraph{Methodology.}
We now evaluate the performance of three representative debiasers,
RW, EOd (as described in \Cref{sec:debiasing_methods}), and NLP (\Cref{alg:nlinprog})
on nine different debiasing experiments as stated in \Cref{tab:datasets},
representing different fairness criteria, data sets and debiasing strategies.
We observe the phenomena in this section when running similar experiments
using the Python toolkits Aequitas \citep{aequitas} and Fairness 360 \citep{aif360},
and have carefully reimplemented the algorithms in our own Julia implementation
(provided in the Supplement) to verify that these effects are not the results of
undiagnosed implementation bugs. We present results from our own implementations,
which corroborate similar findings from the Python codes.

The classifier trained for each experiment is a random forest classifier
estimated using the MIT-licensed \texttt{DecisionTree.jl}~\citep{DecisionTree.jl}
Julia package, which implements the standard classification and regression trees (CART)~\citep{CART}
and random forest algorithms~\citep{Breiman2001}.
While hyperparameter tuning is an important part of developing fair real-world models
\citep{Schelter2019,Perrone2020},
we keep all hyperparameters at the same default values to facilitate comparison across these
varied experiments, eliminating variation due to hyperparameter choice.
Our evaluation criteria are the ratio of out-of-sample fairnesses $\tau/\tau_0$
for the debiased and original classifiers, with $\tau$ as defined in \Cref{def:ratiometric},
and the ratio of out-of-sample accuracies $\gamma/\gamma_0$ respectively.
Unlike many previous studies,
we focus on the \textit{out-of-sample} behavior of the original and debiased classifiers,
and estimate the generalization error
by computing metrics across 100 different train--test splits computed from ten times ten-fold cross-validation (10 CV 10).
Such extensive evaluation is necessary to reduce the error bars on the fairness metrics
$\tau$ to determine if a debiaser had a statistically meaningful treatment effect;
our experiments demonstrating such necessity are detailed in the Supplement.
% Details of the experiments are stated in \Cref{tab:datasets}. Note, that decreasing fairness through additional debiasing can occur due to a multiplicity of reasons, the main two being overfitting of the debiasing method to training data or only partial alignment of debiasing technique and fairness metric. The NLP method furthermore optimizes a non-convex function possibly getting stuck in local minima.
% While we investigate partial debiasing strategies (\Cref{sec:partial_debias}), we do not report results for partially debiased methods in more detail for brevity, and as it is unclear how exactly this hyperparameter would be tuned and selected in a multi-objective fashion.
% Here, only full debiasing is considered.

\begin{table*}[ht]
\centering
% \begin{tabular*}{@{\extracolsep{\fill}}|>{\centering}p{2.5cm}|>{\centering}p{1cm}|>{\centering}p{1cm}|>{\centering}p{2cm}|>{\centering}p{1cm}||>{\centering}p{2cm}
% \scalebox{0.8}{
\begin{tabular}{|c|c|c|c|c|}
\hline
& Data set & Protected class & Fairness metric & Source\tabularnewline
\hline
\textbf{A} & Adult income & sex & PP & \citep{ucimlrepo,Kohavi1996}\tabularnewline
\hline
\textbf{B} & German credit & marital\_status & EFOR & \citep{ucimlrepo}\tabularnewline
\hline
\textbf{C} & Portuguese bank marketing & gender & EFOR & \citep{ucimlrepo}\tabularnewline
\hline
\textbf{D} & COMPAS & race & EFPR & \citep{compas} \tabularnewline
\hline
\textbf{E} & Loan Defaults & sex & EFOR &  \citep{ucimlrepo} \tabularnewline
\hline
% \hline
\textbf{F} & Student Performance & sex & EFNR & \citep{ucimlrepo}\tabularnewline
\hline
\textbf{G} & Communities and crime &  racepctblack & EFPR & \citep{ucimlrepo}\tabularnewline
\hline
\textbf{H} & Framingham Heart Study & male & EFOR &  \citep{fram} \tabularnewline
\hline
\textbf{I} & Medical Expenditure & race & EFOR &  \citep{aif360} \tabularnewline
%\hline
%HMDA Mortgage. & derived\_sex & EFOR & \citep{hmda}\tabularnewline
\hline
\end{tabular}
% }
\caption{List of experiments with data sets and associated fairness metrics used in our benchmarking study of \Cref{sec:benchmark}.}
\label{tab:datasets}
\end{table*}

\begin{figure}
\centering
\includegraphics[width=\textwidth]{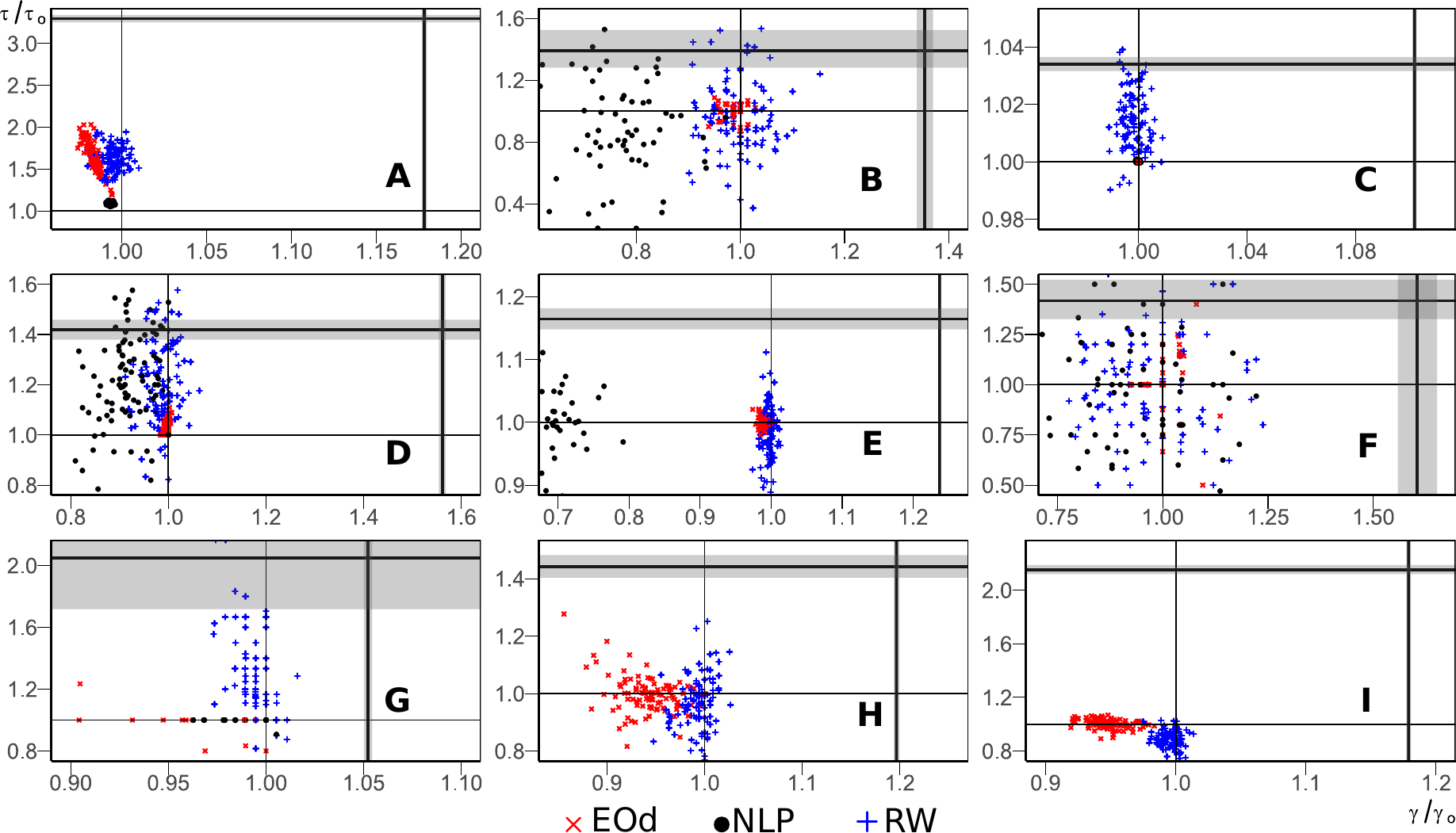}
\vspace{-15pt}
\caption{Plots of fairness ratios $\tau/\tau_0$ (vertical axes) against accuracy ratios $\gamma/\gamma_0$ (horizontal axes) for the experiments of \Cref{sec:benchmark} and \Cref{tab:datasets} using random forest classifiers,
showing that none of the reweighing (RW), equalized odds (EOd), and NLinProg (NLP) debiasers
can consistently debias all the experiments.}
\label{fig:matrix_scatter}
\end{figure}

\paragraph{Results.}%
\label{sec:benchmark_results}
\Cref{fig:matrix_scatter} summarizes the results of our experiments.
Within each subplot and point type, each point corresponds to the exact same classifier type,
debiased the exact same way, but repeated over 100 different train--test splits
arising from ten times ten-fold cross-validation (10 CV 10).
The only variation thus comes from the specific subset of data used for model training,
and the test data for evaluation.
The thick cross-hairs represent the ideal perfect fairness and accuracy,
with the grey regions representing a one standard deviation spread across the folds.
The narrow cross-hairs pinpoint the point where $\gamma = \gamma_0$ and $\tau = \tau_0$, i.e.,
where the debiaser had no treatment effect whatsoever.

Na\"ively, we would expect that $\tau > \tau_0$ and $\gamma \approx \gamma_0$, i.e.,
that the fairness should improve while the accuracy stays roughly constant or perhaps
decreases due to an implicit fairness--accuracy trade-off.
Instead, we see that for Experiments A, E, H, and I,
no debiaser was able to attain the target of maximal fairness.
In fact, some experiments (like NLP in A or EOd in D) show
essentially no change in the fairness and accuracy metrics at all.
More worryingly, nearly all the experiments show large scatter
in the out-of-sample fairness, with many points \textit{below}
the $\tau = \tau_0$ line.
Our results therefore show that not only are debiasers
unable to guarantee fairness out-of-sample,
but even when it can do so for a particular train--test split,
the effect can disappear entirely for different test data.

Experiments A, C, G, H and I also show evidence of an fairness--accuracy
trade-off: as the fairness improves, the accuracy worsens, and the
graphs generally trace out a negative slope.
The satisfiability analysis of \citet{fact}
shows that perfect accuracy and fairness can be attained in theory
for all the experiments; however,
we can understand this effect as arising from change in Bayes rate
due to the additional fairness constraint imposed \citep{fact}.
Nevertheless, we also see evidence of overfitting,
not just in the variance of $\gamma/\gamma_0$,
but also in many points with $\gamma > \gamma_0$,
where debiasing \textit{increased} the accuracy of the classifier,
but not in a robust way.
Our results agree with \citet{Friedler2019},
who showed that debiasing methods are prone to overfit on the training set,
in that debiasing outcomes vary depending on the details of the train/test split,
albeit without an explanation for this phenomenon.
Our results are also consistent with \cite{Schelter2019},
who showed that retuning hyperparameters is necessary to improve generalizability,
also we do not investigate the effect of hyperparameter tuning in our work.

% Our results show considerable variance in the resulting debiased classifiers.
% While some classifiers can achieve essentially perfect fairness, especially through reweighing and NLinProg,
% a considerable number also demonstrate worsened fairness after debiasing.
% Interestingly, reweighing can also lead to \textit{increased} accuracy, which is indicative of overfitting,
% whereas NLinProg does not show such tendencies.
% Finally, EOd shows a much smaller variance in the debiased classifiers,
% % while no classifier attains perfect fairness, the degree to which fairness and accuracy changes is smaller,
% again underscoring the importance of debiasing and measuring with respect to the same measure of fairness.
% % While the small size of this data set exaggerates the variation observed,
% Hence, reliable debiasing requires carefully estimating the intrinsic variability in the fairness metrics.
% % for classifier variance

% What we want to report here:
% - Accuracy drops through debiasing
% - Fairness sometimes even gets worse
% - Overfitting
% Experiment \textbf D is a particularly egregious case where the model becomes more unfair and less accurate, where the accuracy
% halved.
% Our results confirm the trends from previous experiments that
% actually leads to more unfair outcomes in comparison to a simple not-debiased model.
In summary,
\begin{enumerate*}%[1)]
\item
the large variance in fairness metrics necessitate extensive uncertainty quantification to ascertain the treatment effect,
\item
despite controlling for this variance,
fairness can either improve or worsen after debiasing,
and
\item
accuracy usually decreases after debiasing,
sometimes severely so.
\end{enumerate*}
Below in \Cref{sec:learningtheory}, we provide a theoretical analysis of these phenomena in
\Cref{prop:var} in terms of bias-variance trade-off.
In additional experiments in \Cref{sec:partial_debiasing} and the Supplement,
we also demonstrate the somewhat counter-intuitive result that a \textit{partial}
debiasing treatment can actually yield more fair classifiers in practice,
which is the case for 12 out of the 27 combinations of experiment and debiaser.

\section{Convergence of performance--fairness trade-offs}
%Analysis of estimator variance from learning theory}%
\label{sec:learningtheory}

We now present a theoretical analysis of the phenomena we have observed above.
% In this section, we provide some mathematical intuition behind some of the results that we have obtained.
To simplify our approach, we consider the penalized (or dual) version of machine learning problems involving fairness constraints.
% In this simplified setup, which can be easily extended, there are two groups $S_0$ and $S_1$ in the population, characterized by their protected class, $S \in \mathcal{S}$, such that $\forall j \in S_0$, $s_i = 0$ and $\forall j \in S_1$, $s_i = 1$.
% Let $f$ be some model in some model class $\mathcal H$ and $Z = (X, Y)$ be labelled data.
Our starting point is the scalarized optimization program $\lambda \ell + (1-\lambda) \phi$,
% which linearly interpolates between a performance loss $\ell$
% = \ell(f, Z, S)$
% such as the hinge or binomial deviance,
% and an unfairness loss $\phi = \phi(x, x^\prime)$ that vanishes when $x = x^\prime$,
% and $x$ is some metric that depends on protected class $S$ \citep{fact}.
with fairness loss $\phi: \mathbb R^2 \rightarrow \mathbb R^+_0$,
for example, $\phi(x, y) = |x - y|$.
% where in the previous experiments, the fairness loss
% $\phi(x, y) = \tau_h(z)$.
% measuring the discrepancy between a chosen fairness metric $f$ between groups $S=0$ and $S=1$,
The trade-off is parameterized by $\lambda$,
interpolating linearly between considering only fairness ($\lambda=0$)
and only performance ($\lambda=1$).

We want to know how the \textit{empirical} trade-off, as measured on some test set $\mathcal{D}^\star$,
converges to the \textit{true} trade-off, as measured on the true underlying distribution $(Z,S) \sim \mathcal{P}$.
\begin{definition}\label{def:poprisk}
Let
$\ell, \mu: \mathcal H \x{} Z \x{} S \rightarrow \{0, 1\}$
% $ \mu: \mathcal H \x{} Z \x{} S \rightarrow \{0, 1\}$
be indicator functions corresponding to the performance and fairness criteria
such that when the desired criteria are satisfied,
$\mathbb{E}_{(Z,S) \sim \mathcal{P}}(\ell(f,Z,S)) = 0$,
and
$\phi(\tilde z_0, \tilde z_1) = 0$,
where $\tilde z_s = \mathbb{E}_{(Z_{|S\EQ0}) \sim \mathcal{P}}(\mu(f,Z,S))$.
Then, the \textbf{population empirical risk} $L_\P$ for a population $\P$ is
%
% \begin{multline}
% $
\begin{equation}
    L_{\mathcal{P}}(f) = \lambda \mathbb{E}_{(Z,S) \sim \mathcal{P}}(\ell(f,Z,S)) + (1-\lambda) \phi\left(\tilde z_0, \tilde z_1\right).
\end{equation}
    % \label{eq:comboloss}
% $
% \end{multline}
% where $\mu$ %: \mathcal H \x{} (\mathcal X \x{} \mathcal Y) \x{} \mathcal S \rightarrow \mathbb R$
% is an indicator function
% \footnote{This definition can be generalized immediately to multiple dimensions $\mathbb R^n$,
% but we do not consider them further.}
% like misclassification error ($\mu = \mathbf{1}_{\lbrace \hat Y \neq Y \rbrace}$),
\end{definition}

An example of $\ell$ would be misclassification error
$\ell = \mathbf{1}_{\lbrace \hat Y \neq Y \rbrace}$ (the complement of accuracy, $\mathbf{1}_{\lbrace \hat Y \EQ{} Y \rbrace})$,
while an example of $\mu$ would be predictive parity,
$\mu = \mathbf{1}_{\lbrace \hat Y \EQ1 \rbrace}$, corresponding to the fairness constraint
$\Pr(\hat Y\EQ1 | S\EQ1) = \Pr(\hat Y\EQ1 | S\EQ0)$, i.e., demographic parity.
The fairness loss $\phi$ is related to the symmetrized fairness gap $\Delta_h$ defined in \Cref{def:ratiometric},
since we can take $\phi(z_{|S\EQ0}, z_{|S\EQ1}) = |h(z_{|S\EQ0}) - h(z_{|S\EQ1})| = \Delta_h(z)$.
\begin{definition}
The \textbf{sample empirical risk} for a data set $\mathcal D$ is\\
% \scalebox{.9}{\parbox{\columnwidth}{
\begin{equation}
L_{\mathcal{D}}(f) =\lambda l^{(m)}(\mathcal D) + (1-\lambda) \phi\left(l_0^{m_0}(\mathcal D), l_1^{m_1}(\mathcal D)\right),
\end{equation}
% }
%  & = \lambda \mathcal{L}_{\mathcal{D}^\star}(f) + (1-\lambda) \phi \left(M_{{\mathcal{D}^\star},0}(f), M_{{\mathcal{D}^\star},1}(f) \right),
% \label{eq:empiricalrisk}
% \vspace{-30pt}
% }}\\
where $l^{(m)}(\mathcal D) = \sum_{(z,s)\in\mathcal D} \ell(f,z,s) / m$
is the mean empirical performance loss,
$l_s^{(m_s)}(\mathcal D) = \sum_{(z,s')\in\mathcal D: s'=s}\mu(f, z, s) / m_s$
is the mean empirical fairness loss for the subgroup $S=s$,
$m_s = |\{(z, s')\in\mathcal D : s'=s\}|$ is the sample sizes for the group $S=s$,
and $m=m_0+m_1 = \vert \mathcal D \vert $.
\end{definition}

We now derive the limiting distribution of $L_{\mathcal{D}^\star}(f)$ and show that it exhibits some form of bias--variance decomposition.
\begin{theorem}
Let $f: \mathcal X \rightarrow \mathcal Y$ be a classification function and
$\ell$ and $\mu$ be the indicator functions of \Cref{def:poprisk}.
Assume that we have observed $m$ iid samples
$\mathcal D = \{ (Z_j, S_j) : (Z_j, S_j) \sim \mathcal P \}_{j=1}^m$
from a population distribution $\mathcal P$,
the variance of $\ell(f,Z,S)$ is finite,
the fairness penalty function $\phi$ is at least once-differentiable,
and the variance of $\mu(f,Z,S)$ is finite.
Then, the sample empirical loss converges asymptotically to the population empirical loss:
% \begin{equation}
$
    \sqrt{m}\left[ L_{\mathcal{D}}(f) - L_{\mathcal{P}}(f)\right] \xrightarrow[m \rightarrow \infty]{} N\left(0, \mathbb{V}_{\lim}(f)\right),
    % \label{eq:combovar}
$
% \end{equation}
with limiting variance %$\mathbb{V}_{\lim}(f)$
\begin{equation}\begin{aligned}
\mathbb{V}_{\lim}(f) & = \lambda^2 \sum_{s\in \mathcal S} \pi_s (\sigma_s^{\ell})^2
+ \lambda^2 \sum_{s\ne s'} \pi_s\pi_{s'} \left(L_{\mathcal{P},s}(f)-L_{\mathcal{P},s'}(f)\right)^2\\
+ & (1-\lambda)^2 \sum_s k_s^2 \frac{(\sigma_s^{\mu})^2}{\pi_s} + 2\lambda(1-\lambda) \sum_s k_s \Cov_{(z, s')\in\mathcal \mathcal D_s}\left(\ell(f,z,s), \mu(f,z,s)\right),
\label{eq:varianceterms}
\end{aligned}
\end{equation}
where $\Cov$ is the covariance, $s,s'\in\mathcal S$,
$\mathcal D_s = \{(z, s) \in \mathcal D : s'\EQ{}s\} \subseteq \mathcal D$
is the subset of data with protected class membership $S = s$,
$\pi_s = \text{Pr}[S=s]$ is the base rate of the \underline{\smash{protected class}} $S=s$,
$L_{\mathcal{P},s}(f) = \mathbb{E}_{(z, s)\in\mathcal D_s}(\ell(f,z,s))$
is the sample expected loss $\ell$ over $\mathcal D_s$,
$(\sigma_s^{\ell})^2 = \mathbb{V}_{(z, s')\in\mathcal D_s}(\ell(f,z,s))$
is the sample variance of the loss $\ell$ over $\mathcal D_s$,
$M_{\mathcal{P},s}(f)$ and $(\sigma_s^{\mu})^2$ are the analogous
mean and variance for the loss $\mu$,
and $(k_0,k_1)^T = \nabla \phi \left(M_{\mathcal{P},0}(f), M_{\mathcal{P},1}(f) \right)$
is the gradient of $\phi$ at the true value of the fairness function.
\label{prop:var}
\end{theorem}

This result can be proved with repeated use of the central limit theorem,
the delta method and Slutsky's lemma.
The full proof is included in the Supplement.

The first three terms in the limiting variance $\mathbb{V}_{\lim}(f)$ can be interpreted as
\begin{enumerate*}%[1)]
\item the intra-group variance,
\item the (statistical) bias that measures unfairness through the difference in loss for each group $S$, and
\item the variance stemming from the fairness penalty term.
\end{enumerate*}
The last terms grow with $(1-\lambda)^2 k_i^2$, which intuitively captures how sensitivity to fairness constraints leads to increased variance.
Interestingly, these terms are also inversely proportional to the base rates $\pi_s = \Pr(S=s)$, meaning that imbalance in the \underline{\smash{protected class}} increases the variance.

% The last two lines simply reflect the fact that both statistical and fairness components of the trade-off are based on the same sample.

\section{Empirical dependence on protected class imbalance}%
\label{sec:imbalance}

\Cref{prop:var} predicts that the estimation variance depends on protected class imbalance, specifically, that the standard deviation of the estimated classifier $\sigma(\hat f) \sim 1/\Pr(S=1)$ as $\Pr(S=1)$ goes to zero.
We should therefore expect a similar behavior for the fairness metric $\sigma(\tau) \sim 1/\Pr(S=1)$ computed for the estimated classifer.
We now confirm this dependency on a simple synthetic data generating process
that allows us to vary the base rates in both outcome class $\Pr(Y)$
and protected class $\Pr(S)$.
The details of the synthetic data and experimental setup are given in the Supplement.
We use 10 times repeated 10-fold cross-validation on 20,000 data points and report the standard deviation of $\tau_\textsc{EFPR}$ across the replications.
% \begin{figure}
% \centering
% \includegraphics[width=0.99\columnwidth]{figures/imbalance_fairnes.png}
% \caption{Standard deviations obtained for different fractions of target and protected class groups on simulated data. Larger imbalances correspond to higher variance in the estimation of the fairness metric.}
% \label{fig:yzheat}
% \end{figure}

\begin{wrapfigure}{r}{0.5\textwidth}
\centering
\includegraphics[width=0.48\textwidth]{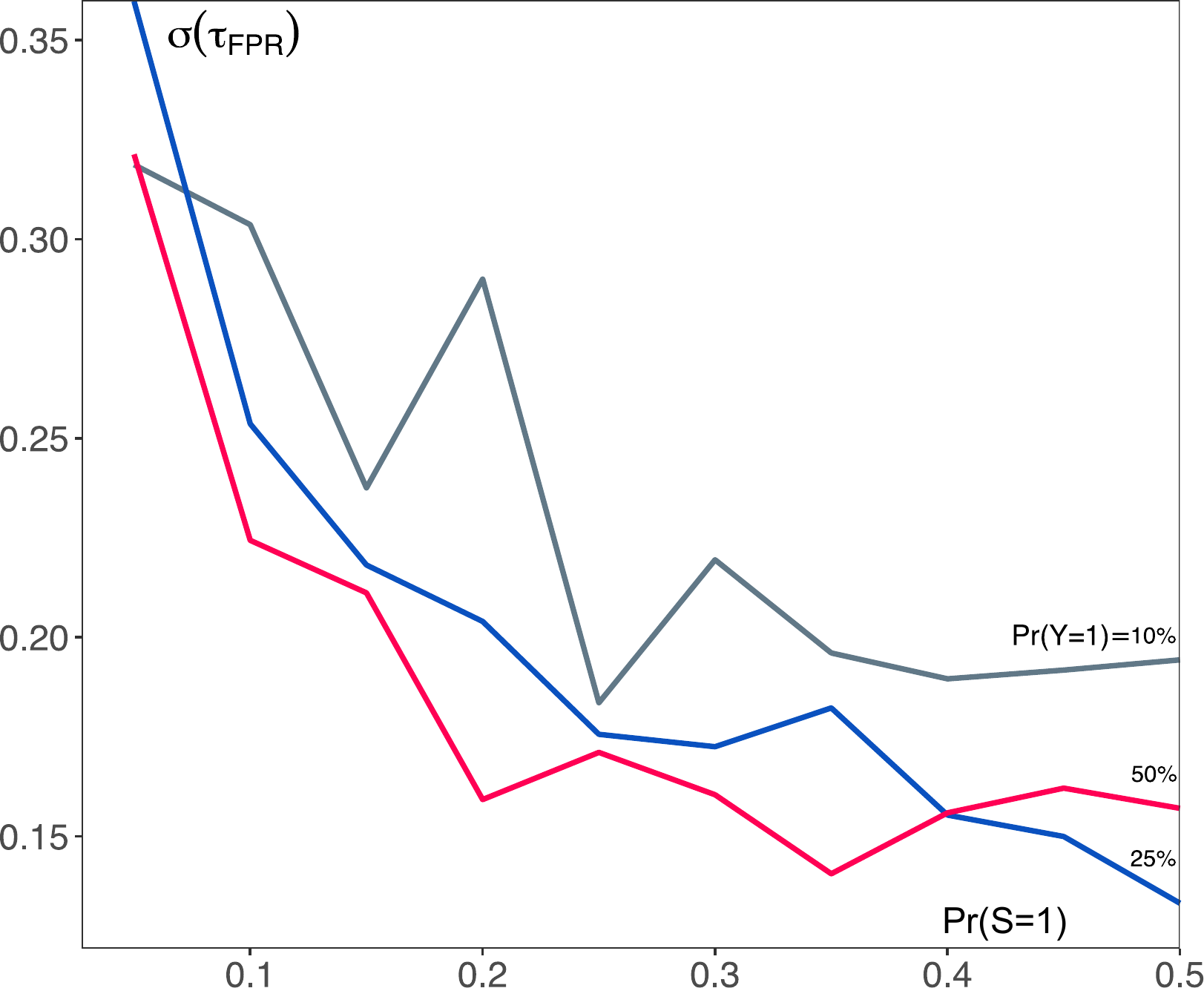}
\caption{Standard deviations for $\tau_\textsc{EFPR}$ estimated via 10-fold CV across different fractions for protected class and positive class. Larger imbalances correspond to higher variance in the estimation of the fairness metric.}
\label{fig:yzheat}
\end{wrapfigure}

\Cref{fig:yzheat} shows how the standard deviation of the fairness metric changes with the base rate $\Pr(S=1)$ for three different values of $\Pr(Y=1)$.
Each curve has the same qualitative shape consistent with an inverse dependence on
$\Pr(S=1)$.
Our results support the theoretical analysis above that the variance in debiasing is strongly affected by class imbalances, both with respect to the class imbalance and the fraction of data points in the two protected attribute groups.

\section{Partial debiasing}%
\label{sec:partial_debiasing}

\Cref{prop:var} implies that when considering the trade-off between
performance and fairness, it is possible to construct a
\textit{variance-minimizing debiaser} that does not perfectly
debias a model, but has better generalization properties.
Minimizing the limiting variance \eqref{eq:varianceterms}
with respect to $\lambda$
will in general not yield a full debiaser $\lambda=1$,
but rather some intermediate debiasing strength.
This observation motivates our introduction of the notion
of \textbf{partial debiasing} in this section.
We will now describe two specific examples of partial debiasing.

\paragraph{Partial reweighing.}
The reweighing pre-processor of \Cref{sec:reweighing} can be easily generalized
to yield a partial debiaser, simply by interpolating between the weight 1 (for $\lambda = 0$) and the weight $w_{h, i}$ for the fairness constraint $h$ in \Cref{def:ratiometric} (for $\lambda=1$).
The simplest such partial reweighing scheme is to simply perform linear interpolation,
% In this paper, we consider only the simplest partial reweighing scheme,
% \vspace{-6pt} % Commented this for NeurIPS
% \begin{equation}
$
w_i = (1-\lambda) 1 + \lambda w_{h, i}
$,
% \label{eq:reweighing_partial}
% \vspace{-6pt} % Commented this for NeurIPS
% \end{equation}
although more exotic interpolation method could also be used.
% as an exemplar of more general partial reweighing schemes that may interpolate among multiple constraints
% and/or in more exotic spaces.
%

\paragraph{Partial post-processing.}
Similarly, for post-processing methods like equalized odds (\Cref{sec:eqodds}) and NLinProg (\Cref{sec:NLinProg}),
we can define a partial debiasing scheme simply by interpolating the flipping probabilities $\Pr(\hat Y^\prime | \hat Y\EQ{}y, S\EQ{}s)$ between 0 and their original values defined previously.
Again for linear interpolation, this corresponds to replacing the flipping probabilities by
$\lambda \Pr(\hat Y^\prime | \hat Y, S)$.

As we show below, we find some surprising and nontrivial behaviors of this simple partial reweighing scheme,
including the result that partial debiasing is in general preferable to full debiasing ($\lambda = 1$)
to produce a low-variance debiased classifier.

\subsection{Empirical evaluation of partial debiasing}

\begin{wrapfigure}{R}{0.35\textwidth}
\centering
\includegraphics[width=0.3\columnwidth]{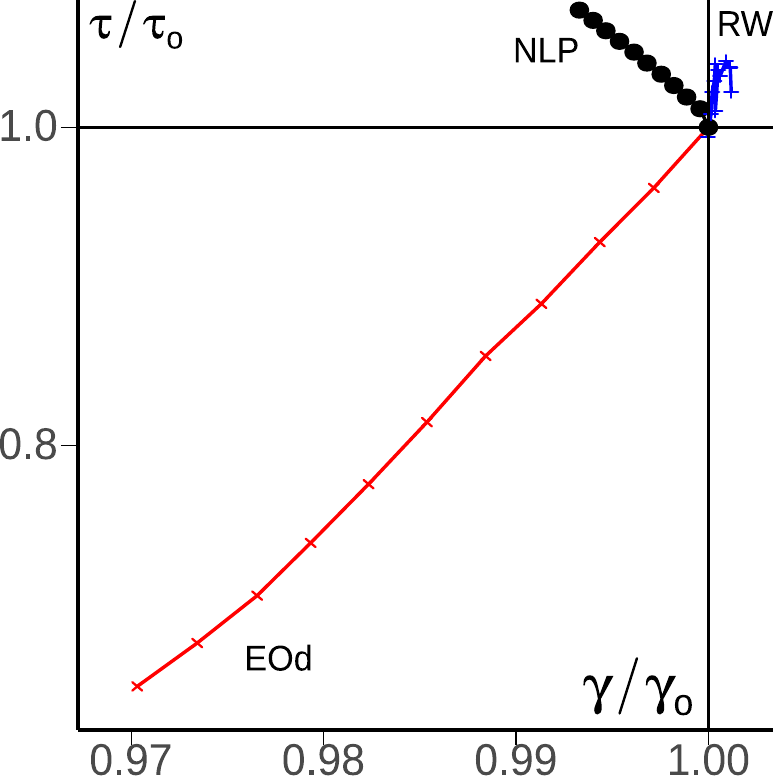}
\vspace{-4pt}
\caption{Accuracy--fairness plot of debiased models derived from a logistic regressor trained for
the Adult income data set and debiased for PP fairness, showing
% the adult income data set
% with accuracy $\gamma_0 = 0.797$ and bias $\tau_{\textrm{PP},0} = 0.543$.
parametric trajectories for increasing the debiasing strength $\lambda$ from 0 to 1
the partial debiasers of \Cref{sec:partial_debiasing}.
% The coordinates are rescaled so that the original classifier has coordinates $(1,1)$,
% and the hypothetical best result is the point of perfect accuracy and fairness with coordinates $(1.254, 1.842)$.
}
\label{fig:adult-ppr}
\end{wrapfigure}

We finish with another follow-up experiment,
reporting out-of-bag metrics after 10-fold cross-validation
and train an initial logistic regression model.
We then debias this same model multiple times with respect to predictive parity (PP) fairness
using three different partial debiasers (\Cref{sec:partial_debiasing}):
partial equalized odds post-processing (EOd),
partial reweighing (RW), and
partial NLinProg (NLP) for accuracy and PP fairness.
\Cref{fig:adult-ppr} shows three different trajectories from parametrically increasing the
debiasing strength from $\lambda = 0$ (no debiasing) to $\lambda = 1$ (full debiasing),
starting from the initial model ($\lambda = 0$) at coordinates $(\gamma/\gamma_0, \tau/\tau_0)= (1, 1)$,
again aggregated across 10CV10 folds.
% The general impossibility result of \citep{fact} demonstrates that accuracy ($\gamma$) and
% PP fairness are compatible in that a trade-off between the two metrics does not exist.
% Therefore, it should be possible to improve a classifier's fairness $\tau$ without compromising accuracy $\gamma$.

As in our earlier experiments in \Cref{sec:benchmark}, we see that the behavior of
the different debiasers are markedly different.
As before, it should be possible in theory to improve a classifier's fairness $\tau$ without compromising accuracy $\gamma$
in this experiment.
On the contrary, we observe that reweighing barely changes the metrics of the model,
whereas EOd steadily leads to worsened PP unfairness and worse accuracy.
The worsened fairness is to be expected, however, since we are debiasing with respect to a different metric that we are measuring.
In contrast, debiasing and measuring the same metric of fairness in NLinProg leads to improved fairness $\tau$,
but at the expense of worsened accuracy $\gamma$.
In this example, none of the debiased models come close to perfect fairness with metric $1/\tau_0 = 1.842$,
implying that the training data and model family simply do not admit a perfectly fair classifier.

\section{Conclusions and outlook}%
\label{sec:conclusions}

We have presented detailed empirical studies throughout the paper (especially \Cref{sec:benchmark})
showing that classifiers treated with debiasing methods
generally suffer from worse out-of-sample generalization behavior,
so much so that the out-of-sample fairness can worsen relative to the original classifier.
We need many test--train--validate splits to make a statistically significant
determination of the treatment effect.
% re needed
% We suggest more care should be taken in assessing the improvement in fairness metric - that means a sufficient number of test/train/validation splits as demonstrated in Figure 1 in the appendix. %  \Cref{fig:cvgen_adult}
As shown in the Supplement, the uncertainty in the fairness metric appears to be usually an order of magnitude larger than that
for accuracy, which could reflect rare protected classes in \Cref{def:ratiometric}.
We showed in \Cref{prop:var} that this increased variance
can be explained by bias--variance trade-off.
To remove the statistical bias
in the classifier that corresponds to discriminatory bias,
we have to impose a fairness constraint, but satisfying that constraint
increases the uncertainty of where the best decision boundary can be drawn,
especially when the baseline model is already carefully estimated with attention paid to out-of-sample
generalization error.
Furthermore, we confirmed empirically in \Cref{sec:imbalance}
that the estimation variance \eqref{eq:varianceterms}
is particularly severe when any of the protected classes is rare,
i.e., when the base rate $\Pr(S)$ approaches zero.
In practice, full debiasing is also not desirable
if the performance of the debiased classifier degrades too much.
We showed in \Cref{sec:partial_debiasing} that the fine-grained
control afforded by partial debiasing
allows us to learn new classifiers that have desirable
out-of-sample fairness properties.
% If we identify the (unfairness) bias of the classifier as a (statistical) bias to be removed,
% and we have a sufficiently flexible model class that admits a representation of a fair classifier,
% we still have to pay the price of having a larger uncertainty in the effectiveness of the debiasing method,
% especially when the baseline model is already carefully estimated with attention paid to out-of-sample
% generalization error.
%  We suggest more care should be taken in assessing the improvement in fairness metric - that means a sufficient number of test/train/validation splits as demonstrated in Figure 1 in the appendix. %  \Cref{fig:cvgen_adult}
% The uncertainty in the fairness metric appears to be usually an order of magnitude larger than that
% for accuracy, which may arise when small denominators exist in \eqref{eq:tau}.

The empirical results, while mostly negative, have motivated the theoretical analysis of \Cref{prop:var},
which gives us detailed insight into the origins of the large variance in classifiers.
In particular, \eqref{eq:varianceterms} states that the variance varies dramatically with
protected class imbalance, which to our knowledge is a new result.
Furthermore, \eqref{eq:varianceterms} suggests that that partial debiasing
can let us find a variance minimizing estimator that, while not applying the full
debiasing treatment, can yield better fairness properties that generalize in practice.
Since it is in general difficult to vary this trade-off parameter $\lambda$ explicitly
to find this estimator, finding practical ways to compute this minimal
variance estimator seems like a promising research direction that could improve the practical utility
of debiasing methods.
Conversely, our results also show that fundamental limits exist to the ability to debias arbitrary models
in a purely black box manner.
Identifying underlying causal connections linking protected classes to features \citep{Zhang2018} may therefore be
a more promising direction for successful mitigation of bias.

Our results signal caution to avoid the risk of fairwashing \citep{Aivodji2019,Anders2020},
in the sense of believing that one is using a fair model resulting
from some debiasing treatment,
when in fact the model is overfit and does not generalize well out-of-sample \citep{Friedler2019}.
Rather than blindly trusting that a debiased classifier is now fair,
our results demonstrate that debiasing treatments need to be carefully tested
in order to verify that the desired fairness properties hold in practice.
\subsection*{Acknowledgements}
% Florian & Bernd
This work has been funded by the German Federal Ministry of Education and Research (BMBF) under Grant No. 01IS18036A. The authors of this work take full responsibilities for its content.
% Sebastian
We also thank our generous funding agencies IQVIA, UNIVERSITY of Auckland, Turing and tools practices, and Microsoft.

\paragraph{Disclaimer}
This paper was prepared for informational purposes in part by the Artificial Intelligence Research group of JPMorgan Chase \& Co and its affiliates (``JP Morgan''), and is not a product of the Research Department of JP Morgan.  JP Morgan makes no representation and warranty whatsoever and disclaims all liability, for the completeness, accuracy or reliability of the information contained herein.  This document is not intended as investment research or investment advice, or a recommendation, offer or solicitation for the purchase or sale of any security, financial instrument, financial product or service, or to be used in any way for evaluating the merits of participating in any transaction, and shall not constitute a solicitation under any jurisdiction or to any person, if such solicitation under such jurisdiction or to such person would be unlawful.

\bibliography{bib}
\bibliographystyle{abbrvnat} % for neurips: Any choice of citation style is acceptable as long as you are consistent

\end{document}

\twocolumn[
\icmltitle{Supplementary Material \\ Debiasing classifiers: is reality at variance with expectation?}
\begin{icmlauthorlist}
\icmlauthor{Ashrya Agrawal}{birla}
\icmlauthor{Florian Pfisterer}{lmu}
\icmlauthor{Bernd Bischl}{lmu}
\icmlauthor{Jiahao Chen}{jpm}
\icmlauthor{Srijan Sood}{jpm}
\icmlauthor{Sameena Shah}{jpm}
\icmlauthor{Francois Buet-Golfouse}{ucl,jpmuk}
\icmlauthor{Bilal A Mateen}{wellcome}
\icmlauthor{Sebastian Vollmer}{warwick,turing}
\end{icmlauthorlist}

\icmlaffiliation{birla}{Birla Institute of Technology and Science, Pilani, India}
\icmlaffiliation{lmu}{Ludwig-Maximilians-University, M\"unich, Germany}
\icmlaffiliation{jpm}{JP Morgan AI Research, New York, New York, USA}
\icmlaffiliation{ucl}{University College London, London, United Kingdom}
\icmlaffiliation{jpmuk}{JP Morgan, London, United Kingdom}
\icmlaffiliation{warwick}{University of Warwick, Warwick, United Kingdom}
\icmlaffiliation{turing}{Alan Turing Institute, London, United Kingdom}
\icmlaffiliation{wellcome}{Wellcome Trust, London, United Kingdom}

\icmlcorrespondingauthor{Jiahao Chen}{jiahao.chen@jpmorgan.com}
\icmlkeywords{Machine Learning, ICML}
\vskip 0.3in
]

% \printAffiliationsAndNotice{} %{\icmlEqualContribution} -- Commmented for NeurIPS

\appendix

\section{Empirical Evaluation}

\subsection{Cross-validation strategies}
We conduct further empirical studies on the simulated datasets used in \citet{Zafar2017-fk} and the real-world Adult dataset \citep{ucimlrepo} in order to show, that this holds for moderate dataset sizes, the additional variance does not depend on stochasticity during training, and that this variance partially stems from the debiasing strategy. \Cref{fig:cvgen_adult} shows estimated $\tau$ for different re-sampling and de-biasing strategies measuring the rate of positive predictions (PPR) and \textit{sex} as a protected class on the Adult dataset. Results depict replications of the full strategy averaged over the respective folds/replications. For fewer folds and non-replicated cross-validation, performance estimates of the different de-biasing strategies clearly overlap despite an actual difference (as evidenced by the medians). A clear separation is only possible for repeated 10-fold cross-validation. This effect can be observed despite the adult dataset containing $50.000$ observations. This poses a problem in practice, when considering that results from e.g. tuning over different debiasing techniques using non-repeated (nested) CV with a low amount of folds might be unreliable.

\begin{figure}[h!tb]
\centering
\includegraphics[width=0.99\columnwidth]{figures/generalization_error_comparison.png}
\caption{Estimated bias $\tau_\textsc{PP}$ across several de-biasing methods and cross-validation strategies on the Adult dataset. This demonstrates that a larger number of folds and replications are required to distinguish methods. }
\label{fig:cvgen_adult}
\end{figure}

\subsection{Variance of the estimation procedure}
In an additional experiment shown in \Cref{fig:syn_zafar} on synthetic data, we aim to separate the effect of variance incurred from different samples in the training data and differing test sets  as well as the effect of stochasticity during the model fitting process by considering a synthetic data scenario. We first fit a model for each debiasing method on $10.000$ datapoints simulated using the approach described in \citet{Zafar2017-ft}. Afterwards, we sample a test-set of size $n=1000$ from the same joint distribution in order to estimate the generalization error of our fairness metric $\tau$ (using False Positive Rates). We repeat the same procedure $30$ times with and without re-fitting the model in each replication in order to assess the additional variance incurred from the stochasticity of the fitting process. In summary, we can observe that: $i)$ the use of debiasing strategies leads to increased variance in the estimate of $\tau$, $ii)$ the variance incurred from the fitting process is negligible in comparison to other factors and $iii)$ that for debasing methods, variance again is larger than differences in medians, underlining the resulting problems for choosing an appropriate debiasing strategy.
\begin{figure}[h!tb]
\centering
\includegraphics[width=0.99\columnwidth]{figures/generalization_error_zafar_joint.png}
\caption{Estimated bias $\tau_{EFPR}$ estimated on multiple test-sets of size $n=1000$ and several de-biasing strategies for data simulated as described in \citep{Zafar2017-ft}. }
\label{fig:syn_zafar}
\end{figure}

\subsection{Summary of experimental results}

\begin{table*}[ht]
\centering
% \scalebox{0.8}{
\begin{tabular}{|l|l|l|l|l|l|l|}
\hline
 & $\gamma_\textrm{EOd}/ \gamma_0$ & $\gamma_\textrm{NLP}/\gamma_0$ & $\gamma_\textrm{RW}/\gamma_0$ & $\tau_\textrm{EOd}/\tau_0$ & $\tau_\textrm{NLP}/\tau_0$ & $\tau_\textrm{RW}/\tau_0$ \\
\hline
\textbf{A} & 0.983[1] & 0.945[1] & \textbf{0.995[2]} & 1.654[58] & 1.686[27] & 1.628[42]\\
\hline
\textbf{B} & 0.760[76] & 0.652[74] & \textbf{0.994[2]} & 0.669[195]* & 0.930[135] & \textbf{1.314[199]}\\
\hline
\textbf{C} & \textbf{0.996[1]} & 0.958[16] & 0.986[9] & 1.035[7] & 1.149[55] & 1.202[55]\\
\hline
\textbf{D} & 0.942[13] & 0.466[13] & \textbf{0.992[5]} & \textbf{0.995[13]} & 0.799[64]* & 0.963[28]\\
\hline
\textbf{E} & 0.991[3] & 0.697[24] & 0.994[15] & 0.999[12] & 0.844[102]* & 0.955[59]\\
\hline
\textbf{F} & 0.991[2] & 0.560[7] & \textbf{0.997[2]} & 0.999[3] & 0.746[49]* & 0.985[14]\\
\hline
\textbf{G} & 0.947[6] & 0.475[17] & \textbf{0.996[2]} & 1.005[11] & 0.994[69] & 0.889[19]*\\
\hline
\textbf{H} & 1.000[0] & 0.999[1] & 0.998[1] & 1.000[0] & 1.007[2] & \textbf{1.014[4]}\\
\hline
\textbf{I} & 0.999[10] & 0.900[32] & 0.968[40] & 0.997[37] & 0.807[97]* & 1.020[74]\\
\hline
\end{tabular}
% }
\caption{Summary of experimental results on random forest classifiers as described in \Cref{sec:benchmark}.
$\gamma_h$ is the accuracy of the classifier after debiasing with method $h$ (0 = no debiasing),
while $\tau_h$ is the fairness measure defined in \Cref{eq:tau} of the classifier after debiasing with method $h$.
Reported values are means computed from 10 times 10-fold cross-validation, with standard errors of the means in brackets
multiplied by a factor of 1000.
% For example, $\gamma_\textrm{EO}/ \gamma_0$ for experiment \textbf A has mean $0.983 \pm 0.001$, and the corresponding
% $\tau_\textrm{RW}/\tau_0$ has mean $1.628 \pm 0.042$.
Numbers in bold denote the best result where statistically significant as described in the main text.
Asterisks denote statistically significant ($>1$ standard error) worsening of fairness after debiasing.
}
\label{tab:resultsb}
\end{table*}

\Cref{tab:resultsb} summarizes the accuracy $\gamma$ and fairness $\tau$ metrics over 10 times 10-fold cross-validation,
and are scaled relative to the performance of the baseline undebiased classifier.
The unscaled metrics can be found in Table 1 in the supplementary material.
% The unscaled metrics are in \Cref{tab:results}.
% for each datasets' associated fairness metric is reported in \ Results are again scaled by dividing by $\gamma_0$ or $\tau_0$ respectively, similar to \Cref{fig:adult-ppr} and \Cref{fig:german-for} to improve readability. Values smaller than $1$ indicate a degradation in comparison to the non-debiased baseline, while values larger then $1$ indicate improvement.
% Reported standard deviations (in brackets) correspond to the standard deviation of a 10-fold CV estimate estimated via replications.
% It is important to consider the standard deviation of $\tau$ in comparison to the differences in mean, even considering computationally demanding CV10 resampling.
Numbers in bold denote statistically significant best results where unambiguous, in the sense that the difference between the best and next best
results differ by at least the square root of the sum of squared standard errors.
In general, reweighing seems to best preserve accuracy, but there seems to be no consensus in which debiasing method
consistently yields the best improvement in fairness.
This lack of definitive result could be addressed with even more extensive cross-validation studies.
%In cases, where $\tau$ is already comparably large, the cost of increased fairness is relatively few points in accuracy.
% The un-centered data for \Cref{tab:resultsb} can be found in \Cref{tab:results} in the supplementary material.
Additionally, there are multiple cases where debiasing worsens the model's unfairness, and most frequently with NLinProg.

Experimental results can be easily reproduced on a single CPU, requiring approximately 35 hours of computation.

\subsection{Details for Imbalance Experiment}

This section describes the experimental setup for the study conducted in Figure $4$ of the main text.
The data generating process for the simple experiment is 
$$y = \beta_0 + \beta_1 x_1 + \beta_2 x_2 + \beta_3 * \mathds{1}_{(S=1)}* x_1 +\beta_4 * \mathds{1}_{(S=1)}* x_2$$
where assignment to classes $0,1$ for a given desired probability of positive instances $p_y$ is conducted via 
$y_{class} := y + \epsilon > quantile(y, p_y)$.
$x_1, x_2$ are sampled from a Multivariate Gaussian $MVN(\mu, \sigma)$ distribution with 
$\mu_1$, $\sigma_1$ = (2, 2),  ((5, 1) (1, 5)) for S = 0 and 
$\mu_2$, $\sigma_2$ = (-2, -2), ((10, 1) (1, 3)) for S = 1 respectively.

For each replication, we sample $25.000$ observations according to $p_y$ and $p_s$ where 
$p_y$, $p_s$ are varied between $\{.05, .1, .15, ... , .5\}$ respectively. 
Standard deviations are computed from $15$ replications of 10-fold CV.
As an ML algorithm, we fit a random forest classifier in it's default settings with subsequent post-processing for equalized odds.

\subsection{Partial Debiasing}%
\label{sec:partial_debiasing}

\begin{figure}
\centering
\includegraphics[width=0.99\columnwidth]{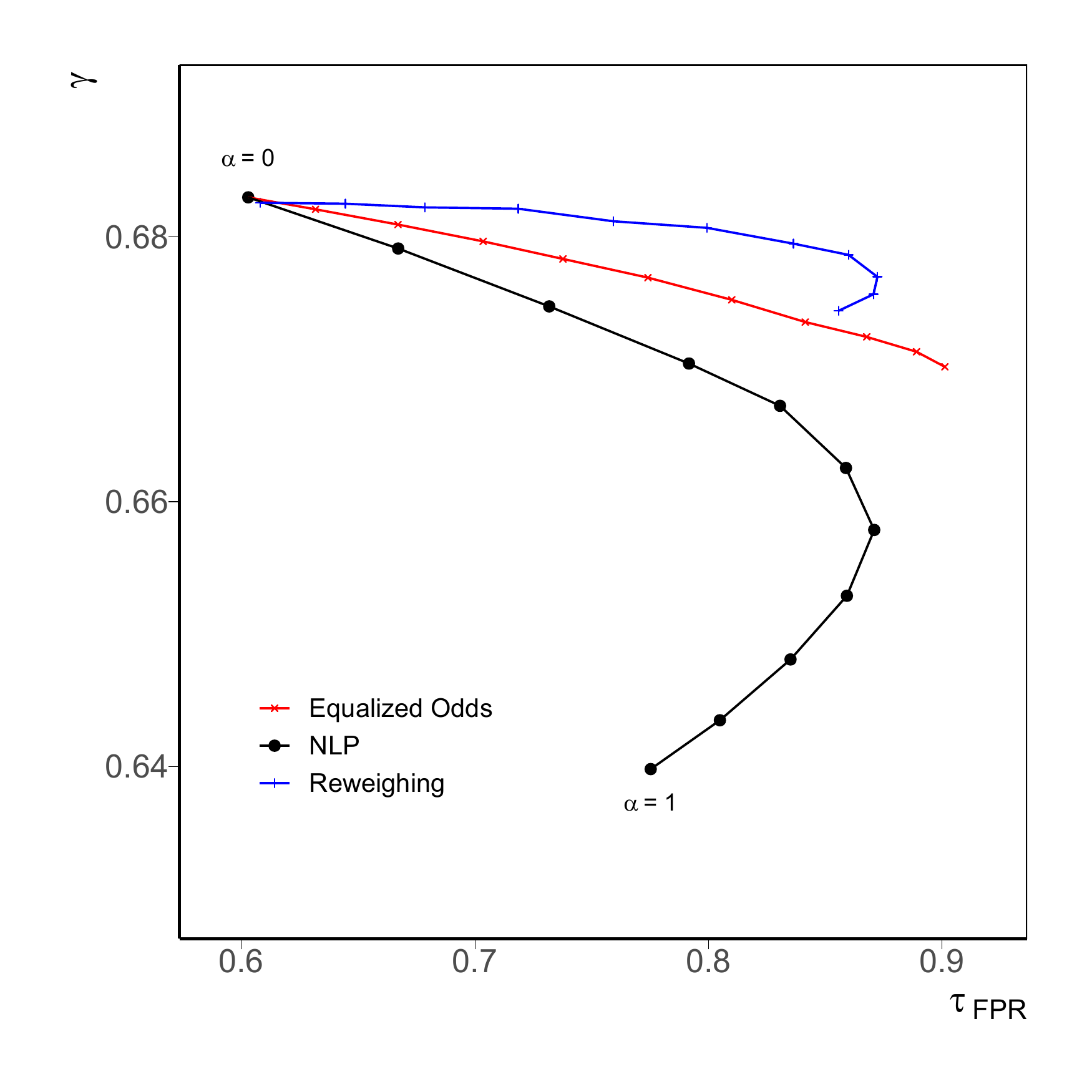}
\caption{Fairness--accuracy trade-off on the COMPAS dataset across several partial debiasing parameters $\alpha$ ( see Equation(7) for a random forest classifier. }
%\eqref{eq:reweighing_partial})
\label{fig:pareto_compas}
\end{figure}

Methods considered in our benchmark exhibit a tunable hyperparameter $\alpha$, allowing to trade-off emphasis on de-biasing against emphasis on accuracy. As the implication of varying this hyperparameter is rarely investigated in literature, we aim to provide additional insights into effects of this trade-off. 
\Cref{fig:pareto_compas} shows the trade-offs available for the investigated methods by varying $\alpha$ between 0 and 1.
As a control, a model without debiasing (None) is added, where the $\alpha$ has no effect and variations in accuracy and $\tau$ therefore are simply a result of stochasticity in the fitting process. Partial debiasing trades off emphasis on accuracy against emphasis on the desired fairness metric, allowing us to obtain a set of various trade-offs. 

\begin{figure}
\centering
\includegraphics[width=0.99\columnwidth]{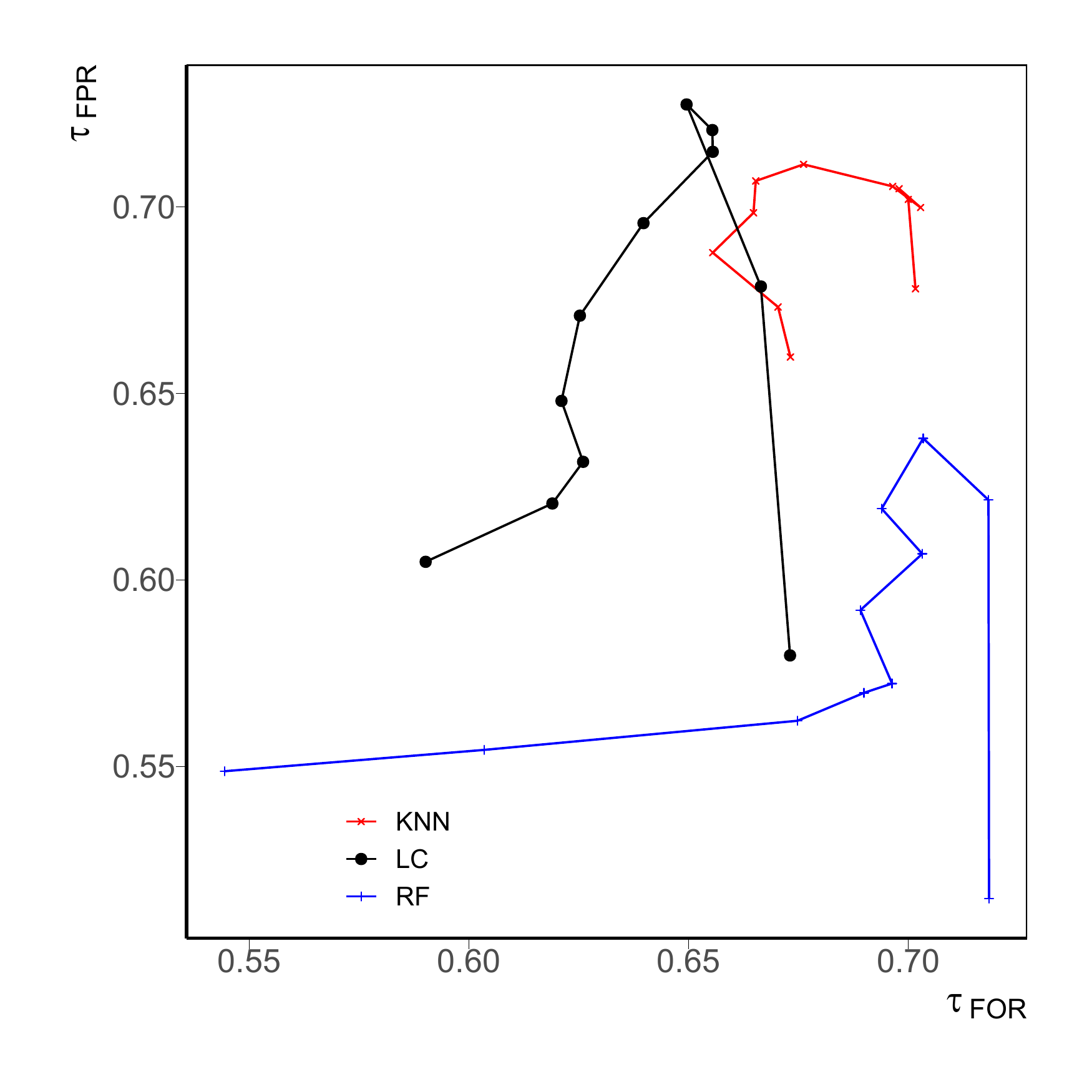}
\caption{Fairness--fairness trade-off on German Dataset when NLinProg algorithm is used to debias with respect to EFOR fairness - across different ranges of $\alpha$, see Equation(7.}
%\eqref{eq:reweighing_partial}
\label{fig:fairness_fairness_german}
\end{figure}

\section{Proof of Proposition 1}
Let us recall some notations:
\begin{multline*}
    L_{\mathcal{P}}(f) = \lambda \mathbb{E}_{(Z,S) \sim \mathcal{P}}[\ell(f,Z,S)] + (1-\lambda)\times \\ \phi\left( \mathbb{E}_{(Z|S=0)}[\mu(f,Z,0)],  \mathbb{E}_{(Z|S=1)}[\mu(f,Z,1)] \right),
    % \label{eq:comboloss2}
\end{multline*}
where $\mu$ %: \mathcal H \times (\mathcal X \times \mathcal Y) \times \mathcal S \rightarrow \mathbb R$
is some indicator function;
like for misclassification error, $\mu = \mathbf{1}_{\lbrace \hat Y \neq Y \rbrace}$,
or
predictive parity, $\mu = \mathbf{1}_{\lbrace \widehat{y}=1 \rbrace}$. Its empirical counterpart is
\begin{multline*}
    L_{\mathcal{D}^\star}(f)  = \frac{\lambda}{m}\sum_{j \in S_0 \cup S_1} \ell(f,z_j,s_j) + \\
                         (1-\lambda)\;\phi\left( \frac{1}{m_0}\sum_{j \in S_0}\mu(f,z_j,0),  
    \frac{1}{m_1}\sum_{j \in S_1}\mu(f,z_j,1) \right),
\end{multline*}
where $m_0$ and $m_1$ are the sample sizes of groups $S=0$ and $S=1$ respectively,
and $m=m_0+m_1 = \vert \mathcal D^\star \vert $. 

Similarly, the theoretical moments can be denoted by
$L_{\mathcal{P},s}(f) = \mathbb{E}_{(Z \vert S =s)}\left[\ell(f,Z,s) \right]$,
$M_{\mathcal{P},s}(f) = \mathbb{E}_{(Z \vert S =s)}\left[\mu(f,Z,s) \right] $, for $s=0,1$, and the corresponding empirical moments as $L_{\mathcal{D^*},s}(f)$ and $M_{\mathcal{D^*},s}(f)$ for $s=0,1$, respectively.
In addition, 
$[\sigma_s^{\ell}]^2 = \mathbb{V}_{Z \vert S=s}[\ell(f,Z,s)]$, $[\sigma_s^{\mu}]^2 = \mathbb{V}_{Z \vert S=s}[\mu(f,Z,s)]$ for $s=0,1$.

Since $\phi$ is differentiable --and under the additional assumption of continuity of its gradient for the sake of simplicity-- applying Taylor's expansion yields
\begin{multline*}
    \phi\left( M_{\mathcal{D^*},0}(f),  
    M_{\mathcal{D^*},1}(f) \right) \\- \phi\left( M_{\mathcal{P},0}(f),  M_{\mathcal{P},1}(f)\right)\\
    = \partial \phi_1 \left(\xi_{m_0},\xi_{m_1} \right) \left(M_{\mathcal{D^*},0}(f) - M_{\mathcal{P},0}(f) \right)\\
    + \partial \phi_2 \left(\xi_{m_0},\xi_{m_1} \right) \left(M_{\mathcal{D^*},1}(f) - M_{\mathcal{P},1}(f)\right),
\end{multline*}
where $\vert\xi_{m_s} -M_{\mathcal{P},s}(f) \vert \leq \vert M_{\mathcal{D^*},s}(f) - M_{\mathcal{P},s}(f) \vert$, for $s=0,1$.

Bringing it all together, it comes
\begin{multline*}
    \sqrt{m}\left(L_{\mathcal{D}^\star}(f) -  L_{\mathcal{P}}(f)\right) =\\ \lambda\sqrt{\frac{m_0}{m}}\sqrt{m_0} \left(L_{\mathcal{D^*},0}(f) -L_{\mathcal{P},0}(f)  \right)\\
    + \lambda\sqrt{\frac{m_1}{m}}\sqrt{m_1} \left(L_{\mathcal{D^*},1}(f) -L_{\mathcal{P},1}(f)  \right)\\
    + \lambda\sqrt{m}\left(\frac{m_0}{m} - \pi_0\right)\left(L_{\mathcal{P},1}(f) -L_{\mathcal{P},0}(f) \right)\\
    +(1-\lambda) \partial \phi_1 \left(\xi_{m_0},\xi_{m_1} \right) \sqrt{\frac{m}{m_0}}\sqrt{m_0}\left(M_{\mathcal{D^*},0}(f) - M_{\mathcal{P},0}(f) \right)\\
    + (1-\lambda)\partial\phi_2 \left(\xi_{m_0},\xi_{m_1} \right) \sqrt{\frac{m}{m_1}}\sqrt{m_1}\left(M_{\mathcal{D^*},1}(f) - M_{\mathcal{P},1}(f) \right).
    \end{multline*}

We note that, almost surely, as $m \rightarrow +\infty$, by the continuous mapping theorem, $\sqrt{\frac{m_0}{m}} \rightarrow \sqrt{\pi_0}$, $\sqrt{\frac{m_1}{m}} \rightarrow \sqrt{1-\pi_0}$, $\partial\phi_k \left(\xi_{m_0},\xi_{m_1} \right) \rightarrow \partial\phi_k \left(M_{\mathcal{P},0}(f),M_{\mathcal{P},1}(f) \right)$ for $k=1,2$. 

All that remains is now to apply the multivariate central limit theorem and Slutsky's lemma to obtain the final result. 

\section{Optimizing for multiple fairness metrics} 

In various fairness problems, picking the debiasing metric can be hard and there might be no clear consensus on it. Generally, we use multiple fairness definitions for a dataset. For example, \citet{corbettcompas} analysis of COMPAS recidivism algorithm uses 3 definitions while \citet{richardcompas} discussion of fairness in COMPAS uses 6 definitions. While optimising a fairness definition, we might end up making other possible definitions of fairness worse.

Results from Proposition 1 can be similarly adapted to account for multiple fairness metrics by allowing for several fairness criteria $\Phi_j$ and multipliers $\lambda_j$ ($j \in \{0,...,J\}$)

\begin{multline}
    L_{\mathcal{P}}(f) = \lambda_0 \mathbb{E}_{(Z,S) \sim \mathcal{P}}[\ell(f,Z,S)] + \\ 
    \sum_j^J \lambda_j
    \phi_{j=1}\left( \mathbb{E}_{(Z|S=0)}[\mu(f,Z,0)],  \mathbb{E}_{(Z|S=1)}[\mu(f,Z,1)] \right)
    \label{eq:2fairloss}
\end{multline}

 with $\sum_0^J \lambda_j = 1$. 
Using the tuneable hyperparameter $\alpha$ described in \Cref{sec:partial_debiasing}, we plot the trade-off between fairness definitions in \Cref{fig:fairness_fairness_german,fig:fairness_fairness_compas}.
In each figure, we measure two different fairness definitions for increasing values of the partial debiasing parameter $\alpha$ across $3$ ML methods: random forest (RF), $k$-nearest neighbours (KNN) and logistic regression (LR). 

\begin{figure}[t]
\centering
\includegraphics[width=0.99\columnwidth]{figures/COMPAS-pareto.pdf}
\caption{Fairness--fairness trade-off on COMPAS Dataset when NLinProg algorithm is used to debias both EFPR and EFNR fairness - across different ranges of $\alpha$, see Equation(7).}
\label{fig:fairness_fairness_compas}
\end{figure}

This allows us to make several interesting observations: Some measures like EFOR and EFPR in \Cref{fig:fairness_fairness_german}, at least for larger values of $\tau$ need to be traded off against each other. Further improvement in one direction leads to decreasing performance for the other measure. The same results hold, when two fairness metrics are jointly optimized (c.f.\ \Cref{fig:fairness_fairness_compas}). Furthermore, for some settings (RF and KNN in \Cref{fig:fairness_fairness_compas}), jointly optimizing two measures (FPR and FNR) without sacrificing one seems possible.

\section{Bias-Variance analysis with a na\"ive post-processing algorithm}

Using the Bias-Variance-Noise decomposition provided by \cite{bias_variance}, we perform the following analysis.

In our analysis, 0 denotes the privileged group, while 1 denotes the unprivileged group.
$\beta_a$ denotes the base rate for group a. p denotes the accuracy of base classifier. We consider L=L1 loss in analysis.
For ease of calculations, we assume that both races exist in same proportion in train and test set.

As analysis with Reweighing, Equalized and NLP is complex, we use a rather simpler na\"ive fairness algorithm described as following:

\[
    F(y, a) = 
        \begin{cases}
        \text{y,}   &\quad\text{with probability }(1-\alpha) \\
        \text{0,}   &\quad\text{with probability }(\frac{\alpha}{2} \left ( 1 - \beta_a \right)) \\
        \text{1,}   &\quad\text{with probability }(\frac{\alpha}{2} \beta_a) \\
        \end{cases}
\]

Assuming randomized features X, and the protected attribute value a, we get:
\\
The main prediction as:
$$\Tilde{y}(a) = \beta_{a} \left (1 - \frac{\alpha}{2} \right)$$
\\
The Bayes optimal prediction is
$$y^{*}(a) = \beta_{a}$$

Making the i.i.d assumption and assuming that fraction of positive predictions by model is in same proportion as train set i.e. $\beta_{a}$.
% We also assume that $\beta_{a} < \frac{1}{2}$
\begin{equation}
\label{eqn:Ba}
    B_{a} = \alpha \frac{\beta_a}{2}
\end{equation}

% Now, we assume difference in accuracy ($p_0 - p_1$) of base classifier over the classes to be insignificant in comparison to difference in base rates, $\beta_0-\beta_1$.
% Let $$k = 4 \left (\frac{1}{2} - \beta_{a} \right) \left (p -\frac{1}{2} \right)$$

\begin{equation}
\label{eqn:Va}
V_{a} =     \left (1 - \frac{\alpha}{2} \right) \left ( \left (1 - \frac{\alpha}{2} \right) (\beta_a - \beta_a^2) + \beta_a \right)
\end{equation}

\begin{equation}
\label{eqn:Na}
N_{a} = 2\beta_a(1-\beta_a)
\end{equation}

As defined by \cite{bias_variance}, the discrimination level $\Gamma$ admits decomposition of the following form
$$\Gamma = |(N_0 - N_1) + (B_0 - B_1) + (V_0 - V_1)|$$

Using equation \ref{eqn:Na}, we write the first term in decomposition as 
$$N_0-N_1 = 2(\beta_0 - \beta_1)(1- (\beta_0 - \beta_1))$$

Using equation \ref{eqn:Ba}, we can write the second term (in decomposition) denoting \textbf{Bias term} as
\begin{equation}
    \label{eqn:B}
    B_0-B_1 = \frac{\alpha}{2}(\beta_0-\beta_1)
\end{equation}

Note that the second term in equation \ref{eqn:B} is positive as $(\beta_0-\beta_1)$ is always positive. Magnitude of the bias term increases as we increase the partial debiasing parameter $\alpha$.\\
\\
Let c denote $\beta_0+\beta_1$. Using equation \ref{eqn:Va}, we can write the third term (in decomposition) denoting variance term as follows:
\begin{equation}
\label{eqn:V}
    V_0-V_1 = \frac{1}{2}(\beta_0-\beta_1) \left (\frac{1}{2} - c \right) (\alpha - 2) \left (\alpha - 4\frac{c-1}{2c-1} \right)  \\
\end{equation}

Note that the third term in the RHS of equation \ref{eqn:V} is a quadratic equation. The equation presents following interesting possibilities depending on exact value of c = $\beta_0+\beta_1$:
\begin{itemize}
    \item  There is a maxima between $\alpha=0$ and $\alpha=1$ at $2 - \frac{1}{2c - 1}$ for $c \in \left (\frac{3}{4}, 1\right)$
    \item Monotonically decreasing with alpha between $\alpha=0$ and $\alpha=1$
    \item Monotonically increasing with alpha between $\alpha=0$ and $\alpha=1$
\end{itemize}

\section{New Figures}

\begin{figure}
    \centering
    \includegraphics[width=0.9\columnwidth]{figures/germanFOR-histogram.pdf}
    \caption{New Figure 3 - German Dataset - FOR}
    \label{fig:my_label}
\end{figure}

\def\sparklineheight ex{30pt}
\colorlet{lightblue}{white!70!blue}

\begin{tabular}{l|l|l|l|l|l|l}
\hline
Data set & $\gamma_{EO}$ & $\gamma_{LP}$ & $\gamma_{RW}$ & $\tau_{EO}$ & $\tau_{LP}$ & $\tau_{RW}$\\
\hline
Adult & \begin{sparkline}{10}
\spark 0 0.780869565217391 0.1 0.777391304347826 0.2 0.775652173913043 0.3 0.773913043478261 0.4 0.770434782608696 0.5 0.768695652173913 0.6 0.765217391304348 0.7 0.763478260869565 0.8 0.76 0.9 0.758260869565217 1 0.754782608695652 /
\sparkdot 0 0.780452173913043 black 
\sparkdot 0.1 0.778104347826087 lightblue 
\sparkdot 0.2 0.775652173913043 lightblue 
\sparkdot 0.3 0.773252173913044 lightblue 
\sparkdot 0.4 0.770747826086957 lightblue 
\sparkdot 0.5 0.768104347826087 lightblue 
\sparkdot 0.6 0.765634782608696 lightblue 
\sparkdot 0.7 0.763113043478261 lightblue 
\sparkdot 0.8 0.760608695652174 lightblue 
\sparkdot 0.9 0.758104347826087 lightblue 
\sparkdot 1 0.755547826086956 blue 
\end{sparkline} & \begin{sparkline}{10}
\spark 0 0.707826086956522 0.1 0.707826086956522 0.2 0.707826086956522 0.3 0.706086956521739 0.4 0.704347826086957 0.5 0.704347826086957 0.6 0.702608695652174 0.7 0.702608695652174 0.8 0.700869565217391 0.9 0.700869565217391 1 0.699130434782609 /
\sparkdot 0 0.708539130434783 black 
\sparkdot 0.1 0.708069565217391 lightblue 
\sparkdot 0.2 0.70704347826087 lightblue 
\sparkdot 0.3 0.706104347826087 lightblue 
\sparkdot 0.4 0.705130434782609 lightblue 
\sparkdot 0.5 0.704069565217391 lightblue 
\sparkdot 0.6 0.703217391304348 lightblue 
\sparkdot 0.7 0.702139130434783 lightblue 
\sparkdot 0.8 0.70104347826087 lightblue 
\sparkdot 0.9 0.700121739130435 lightblue 
\sparkdot 1 0.699130434782609 blue 
\end{sparkline} & \begin{sparkline}{10}
\spark 0 0.775652173913043 0.1 0.775652173913043 0.2 0.775652173913043 0.3 0.775652173913043 0.4 0.773913043478261 0.5 0.773913043478261 0.6 0.773913043478261 0.7 0.773913043478261 0.8 0.773913043478261 0.9 0.772173913043478 1 0.772173913043478 /
\sparkdot 0 0.77535652173913 red 
\sparkdot 0.1 0.776278260869565 black 
\sparkdot 0.2 0.776086956521739 lightblue 
\sparkdot 0.3 0.774834782608696 lightblue 
\sparkdot 0.4 0.774678260869565 lightblue 
\sparkdot 0.5 0.774660869565217 lightblue 
\sparkdot 0.6 0.773930434782609 lightblue 
\sparkdot 0.7 0.774330434782609 lightblue 
\sparkdot 0.8 0.773217391304348 lightblue 
\sparkdot 0.9 0.772573913043478 lightblue 
\sparkdot 1 0.772869565217391 blue 
\end{sparkline} & \begin{sparkline}{10}
\spark 0 -0.168695652173913 0.1 -0.135652173913044 0.2 -0.100869565217391 0.3 -0.0678260869565218 0.4 -0.0330434782608696 0.5 0.00347826086956522 0.6 0.0365217391304347 0.7 0.0713043478260869 0.8 0.106086956521739 0.9 0.140869565217391 1 0.177391304347826 /
\sparkdot 0 -0.168069565217391 red 
\sparkdot 0.1 -0.136313043478261 lightblue 
\sparkdot 0.2 -0.101739130434783 lightblue 
\sparkdot 0.3 -0.0672521739130436 lightblue 
\sparkdot 0.4 -0.0323652173913044 lightblue 
\sparkdot 0.5 0.00266086956521735 lightblue 
\sparkdot 0.6 0.0371652173913044 lightblue 
\sparkdot 0.7 0.0719652173913043 lightblue 
\sparkdot 0.8 0.106591304347826 lightblue 
\sparkdot 0.9 0.141617391304348 lightblue 
\sparkdot 1 0.176573913043478 black 
\end{sparkline} & \begin{sparkline}{10}
\spark 0 0.144347826086956 0.1 0.147826086956522 0.2 0.15304347826087 0.3 0.158260869565217 0.4 0.163478260869565 0.5 0.168695652173913 0.6 0.173913043478261 0.7 0.179130434782609 0.8 0.184347826086957 0.9 0.189565217391304 1 0.19304347826087 /
\sparkdot 0 0.144504347826087 red 
\sparkdot 0.1 0.147513043478261 lightblue 
\sparkdot 0.2 0.152678260869565 lightblue 
\sparkdot 0.3 0.158069565217391 lightblue 
\sparkdot 0.4 0.163130434782609 lightblue 
\sparkdot 0.5 0.16815652173913 lightblue 
\sparkdot 0.6 0.173913043478261 lightblue 
\sparkdot 0.7 0.179113043478261 lightblue 
\sparkdot 0.8 0.184017391304348 lightblue 
\sparkdot 0.9 0.18895652173913 lightblue 
\sparkdot 1 0.193721739130435 black 
\end{sparkline} & \begin{sparkline}{10}
\spark 0 -0.165217391304348 0.1 -0.125217391304348 0.2 -0.0956521739130436 0.3 -0.0608695652173914 0.4 -0.0156521739130435 0.5 0.00695652173913044 0.6 0.0365217391304347 0.7 0.0730434782608695 0.8 0.100869565217391 0.9 0.135652173913043 1 0.163478260869565 /
\sparkdot 0 -0.165234782608696 red 
\sparkdot 0.1 -0.124765217391304 lightblue 
\sparkdot 0.2 -0.0953217391304348 lightblue 
\sparkdot 0.3 -0.0612521739130435 lightblue 
\sparkdot 0.4 -0.0157913043478261 lightblue 
\sparkdot 0.5 0.00718260869565212 lightblue 
\sparkdot 0.6 0.036695652173913 lightblue 
\sparkdot 0.7 0.0728173913043478 lightblue 
\sparkdot 0.8 0.10104347826087 lightblue 
\sparkdot 0.9 0.135547826086956 lightblue 
\sparkdot 1 0.163095652173913 black 
\end{sparkline}\\
\hline
Communities\&Crime & \begin{sparkline}{10}
\spark 0 0.956521739130435 0.1 0.921739130434783 0.2 0.881739130434783 0.3 0.841739130434783 0.4 0.801739130434783 0.5 0.761739130434783 0.6 0.72 0.7 0.68 0.8 0.641739130434783 0.9 0.601739130434783 1 0.561739130434783 /
\sparkdot 0 0.957130434782609 black 
\sparkdot 0.1 0.92144347826087 lightblue 
\sparkdot 0.2 0.88184347826087 lightblue 
\sparkdot 0.3 0.841791304347826 lightblue 
\sparkdot 0.4 0.802104347826087 lightblue 
\sparkdot 0.5 0.761182608695652 lightblue 
\sparkdot 0.6 0.7208 lightblue 
\sparkdot 0.7 0.680591304347826 lightblue 
\sparkdot 0.8 0.641582608695652 lightblue 
\sparkdot 0.9 0.601634782608696 lightblue 
\sparkdot 1 0.560886956521739 blue 
\end{sparkline} & \begin{sparkline}{10}
\spark 0 0.95304347826087 0.1 0.897391304347826 0.2 0.84 0.3 0.782608695652174 0.4 0.725217391304348 0.5 0.669565217391304 0.6 0.612173913043478 0.7 0.554782608695652 0.8 0.497391304347826 0.9 0.44 1 0.382608695652174 /
\sparkdot 0 0.952973913043478 black 
\sparkdot 0.1 0.897565217391304 lightblue 
\sparkdot 0.2 0.84 lightblue 
\sparkdot 0.3 0.783408695652174 lightblue 
\sparkdot 0.4 0.725878260869565 lightblue 
\sparkdot 0.5 0.669130434782609 lightblue 
\sparkdot 0.6 0.612504347826087 lightblue 
\sparkdot 0.7 0.554382608695652 lightblue 
\sparkdot 0.8 0.497860869565217 lightblue 
\sparkdot 0.9 0.439878260869565 lightblue 
\sparkdot 1 0.3824 blue 
\end{sparkline} & \begin{sparkline}{10}
\spark 0 0.95304347826087 0.1 0.954782608695652 0.2 0.954782608695652 0.3 0.954782608695652 0.4 0.95304347826087 0.5 0.951304347826087 0.6 0.951304347826087 0.7 0.949565217391304 0.8 0.949565217391304 0.9 0.947826086956522 1 0.946086956521739 /
\sparkdot 0 0.953478260869565 red 
\sparkdot 0.1 0.954504347826087 lightblue 
\sparkdot 0.2 0.954782608695652 black 
\sparkdot 0.3 0.954608695652174 lightblue 
\sparkdot 0.4 0.952765217391304 lightblue 
\sparkdot 0.5 0.951634782608696 lightblue 
\sparkdot 0.6 0.951026086956522 lightblue 
\sparkdot 0.7 0.950417391304348 lightblue 
\sparkdot 0.8 0.950069565217391 lightblue 
\sparkdot 0.9 0.948486956521739 lightblue 
\sparkdot 1 0.9464 blue 
\end{sparkline} & \begin{sparkline}{10}
\spark 0 0.15304347826087 0.1 0.130434782608696 0.2 0.102608695652174 0.3 0.0678260869565217 0.4 0.0278260869565217 0.5 0.0260869565217391 0.6 0.0208695652173912 0.7 -0.0260869565217392 0.8 -0.0904347826086957 0.9 -0.132173913043478 1 -0.142608695652174 /
\sparkdot 0 0.153721739130435 black 
\sparkdot 0.1 0.131060869565217 lightblue 
\sparkdot 0.2 0.103095652173913 lightblue 
\sparkdot 0.3 0.0685043478260869 lightblue 
\sparkdot 0.4 0.0276521739130434 lightblue 
\sparkdot 0.5 0.0263130434782608 lightblue 
\sparkdot 0.6 0.020191304347826 lightblue 
\sparkdot 0.7 -0.0258782608695652 lightblue 
\sparkdot 0.8 -0.0896695652173913 lightblue 
\sparkdot 0.9 -0.132730434782609 lightblue 
\sparkdot 1 -0.142730434782609 blue 
\end{sparkline} & \begin{sparkline}{10}
\spark 0 0.179130434782609 0.1 0.144347826086956 0.2 0.116521739130435 0.3 0.0765217391304348 0.4 0.0347826086956521 0.5 0.0226086956521738 0.6 -0.0156521739130435 0.7 -0.0521739130434783 0.8 -0.0486956521739131 0.9 -0.0382608695652174 1 0.0852173913043478 /
\sparkdot 0 0.179652173913043 black 
\sparkdot 0.1 0.144017391304348 lightblue 
\sparkdot 0.2 0.117113043478261 lightblue 
\sparkdot 0.3 0.0771130434782609 lightblue 
\sparkdot 0.4 0.0345565217391304 lightblue 
\sparkdot 0.5 0.0217565217391304 lightblue 
\sparkdot 0.6 -0.0162434782608696 lightblue 
\sparkdot 0.7 -0.0523304347826087 lightblue 
\sparkdot 0.8 -0.0489391304347826 lightblue 
\sparkdot 0.9 -0.0378956521739131 lightblue 
\sparkdot 1 0.086 blue 
\end{sparkline} & \begin{sparkline}{10}
\spark 0 0.172173913043478 0.1 0.163478260869565 0.2 0.151304347826087 0.3 0.194782608695652 0.4 0.215652173913044 0.5 0.285217391304348 0.6 0.297391304347826 0.7 0.300869565217391 0.8 0.361739130434783 0.9 0.386086956521739 1 0.403478260869565 /
\sparkdot 0 0.172017391304348 red 
\sparkdot 0.1 0.163808695652174 lightblue 
\sparkdot 0.2 0.150626086956522 lightblue 
\sparkdot 0.3 0.194573913043478 lightblue 
\sparkdot 0.4 0.214817391304348 lightblue 
\sparkdot 0.5 0.284452173913043 lightblue 
\sparkdot 0.6 0.29784347826087 lightblue 
\sparkdot 0.7 0.300417391304348 lightblue 
\sparkdot 0.8 0.361652173913043 lightblue 
\sparkdot 0.9 0.385234782608696 lightblue 
\sparkdot 1 0.404034782608696 black 
\end{sparkline}\\
\hline
COMPAS & \begin{sparkline}{10}
\spark 0 0.419130434782609 0.1 0.419130434782609 0.2 0.417391304347826 0.3 0.417391304347826 0.4 0.417391304347826 0.5 0.417391304347826 0.6 0.415652173913043 0.7 0.415652173913043 0.8 0.415652173913043 0.9 0.413913043478261 1 0.413913043478261 /
\sparkdot 0 0.418434782608696 black 
\sparkdot 0.1 0.418417391304348 lightblue 
\sparkdot 0.2 0.4176 lightblue 
\sparkdot 0.3 0.417652173913043 lightblue 
\sparkdot 0.4 0.417269565217391 lightblue 
\sparkdot 0.5 0.416886956521739 lightblue 
\sparkdot 0.6 0.416260869565217 lightblue 
\sparkdot 0.7 0.415704347826087 lightblue 
\sparkdot 0.8 0.415286956521739 lightblue 
\sparkdot 0.9 0.414730434782609 lightblue 
\sparkdot 1 0.414 blue 
\end{sparkline} & \begin{sparkline}{10}
\spark 0 0.44695652173913 0.1 0.44 0.2 0.43304347826087 0.3 0.424347826086957 0.4 0.417391304347826 0.5 0.410434782608696 0.6 0.401739130434783 0.7 0.394782608695652 0.8 0.387826086956522 0.9 0.379130434782609 1 0.372173913043478 /
\sparkdot 0 0.447495652173913 black 
\sparkdot 0.1 0.440382608695652 lightblue 
\sparkdot 0.2 0.432921739130435 lightblue 
\sparkdot 0.3 0.424817391304348 lightblue 
\sparkdot 0.4 0.417582608695652 lightblue 
\sparkdot 0.5 0.409965217391304 lightblue 
\sparkdot 0.6 0.402591304347826 lightblue 
\sparkdot 0.7 0.395095652173913 lightblue 
\sparkdot 0.8 0.387078260869565 lightblue 
\sparkdot 0.9 0.379321739130435 lightblue 
\sparkdot 1 0.371947826086957 blue 
\end{sparkline} & \begin{sparkline}{10}
\spark 0 0.40695652173913 0.1 0.408695652173913 0.2 0.405217391304348 0.3 0.403478260869565 0.4 0.405217391304348 0.5 0.403478260869565 0.6 0.403478260869565 0.7 0.405217391304348 0.8 0.405217391304348 0.9 0.405217391304348 1 0.403478260869565 /
\sparkdot 0 0.407513043478261 red 
\sparkdot 0.1 0.408782608695652 black 
\sparkdot 0.2 0.404921739130435 lightblue 
\sparkdot 0.3 0.403704347826087 lightblue 
\sparkdot 0.4 0.404452173913044 lightblue 
\sparkdot 0.5 0.403756521739131 lightblue 
\sparkdot 0.6 0.403913043478261 lightblue 
\sparkdot 0.7 0.405234782608696 lightblue 
\sparkdot 0.8 0.405078260869565 lightblue 
\sparkdot 0.9 0.404869565217391 lightblue 
\sparkdot 1 0.402626086956522 blue 
\end{sparkline} & \begin{sparkline}{10}
\spark 0 0.530434782608696 0.1 0.530434782608696 0.2 0.535652173913043 0.3 0.540869565217391 0.4 0.546086956521739 0.5 0.55304347826087 0.6 0.556521739130435 0.7 0.56 0.8 0.565217391304348 0.9 0.570434782608696 1 0.573913043478261 /
\sparkdot 0 0.530852173913043 red 
\sparkdot 0.1 0.530852173913043 lightblue 
\sparkdot 0.2 0.535269565217391 lightblue 
\sparkdot 0.3 0.540991304347826 lightblue 
\sparkdot 0.4 0.546817391304348 lightblue 
\sparkdot 0.5 0.552382608695652 lightblue 
\sparkdot 0.6 0.556539130434783 lightblue 
\sparkdot 0.7 0.560469565217391 lightblue 
\sparkdot 0.8 0.565965217391304 lightblue 
\sparkdot 0.9 0.570730434782609 lightblue 
\sparkdot 1 0.574330434782609 black 
\end{sparkline} & \begin{sparkline}{10}
\spark 0 0.467826086956522 0.1 0.492173913043478 0.2 0.52 0.3 0.549565217391304 0.4 0.579130434782609 0.5 0.610434782608696 0.6 0.645217391304348 0.7 0.674782608695652 0.8 0.707826086956522 0.9 0.723478260869565 1 0.71304347826087 /
\sparkdot 0 0.467078260869565 red 
\sparkdot 0.1 0.492208695652174 lightblue 
\sparkdot 0.2 0.520278260869565 lightblue 
\sparkdot 0.3 0.549617391304348 lightblue 
\sparkdot 0.4 0.578330434782609 lightblue 
\sparkdot 0.5 0.61064347826087 lightblue 
\sparkdot 0.6 0.645947826086957 lightblue 
\sparkdot 0.7 0.675165217391304 lightblue 
\sparkdot 0.8 0.707913043478261 lightblue 
\sparkdot 0.9 0.723408695652174 black 
\sparkdot 1 0.712260869565217 blue 
\end{sparkline} & \begin{sparkline}{10}
\spark 0 0.556521739130435 0.1 0.570434782608696 0.2 0.594782608695652 0.3 0.659130434782609 0.4 0.666086956521739 0.5 0.681739130434783 0.6 0.699130434782609 0.7 0.740869565217391 0.8 0.747826086956522 0.9 0.772173913043478 1 0.777391304347826 /
\sparkdot 0 0.555982608695652 red 
\sparkdot 0.1 0.570486956521739 lightblue 
\sparkdot 0.2 0.594765217391304 lightblue 
\sparkdot 0.3 0.659426086956522 lightblue 
\sparkdot 0.4 0.666330434782609 lightblue 
\sparkdot 0.5 0.680973913043478 lightblue 
\sparkdot 0.6 0.699878260869565 lightblue 
\sparkdot 0.7 0.740295652173913 lightblue 
\sparkdot 0.8 0.747947826086957 lightblue 
\sparkdot 0.9 0.772904347826087 lightblue 
\sparkdot 1 0.776782608695652 black 
\end{sparkline}\\
\hline
Framingham & \begin{sparkline}{10}
\spark 0 0.758260869565217 0.1 0.751304347826087 0.2 0.740869565217391 0.3 0.733913043478261 0.4 0.725217391304348 0.5 0.716521739130435 0.6 0.707826086956522 0.7 0.699130434782609 0.8 0.690434782608696 0.9 0.681739130434783 1 0.674782608695652 /
\sparkdot 0 0.757704347826087 black 
\sparkdot 0.1 0.750660869565217 lightblue 
\sparkdot 0.2 0.741669565217391 lightblue 
\sparkdot 0.3 0.733182608695652 lightblue 
\sparkdot 0.4 0.724852173913044 lightblue 
\sparkdot 0.5 0.71615652173913 lightblue 
\sparkdot 0.6 0.707913043478261 lightblue 
\sparkdot 0.7 0.698852173913043 lightblue 
\sparkdot 0.8 0.69064347826087 lightblue 
\sparkdot 0.9 0.6824 lightblue 
\sparkdot 1 0.67415652173913 blue 
\end{sparkline} & \begin{sparkline}{10}
\spark 0 0.76 0.1 0.683478260869565 0.2 0.60695652173913 0.3 0.528695652173913 0.4 0.452173913043478 0.5 0.372173913043478 0.6 0.293913043478261 0.7 0.215652173913044 0.8 0.139130434782609 0.9 0.0608695652173913 1 -0.0191304347826087 /
\sparkdot 0 0.759652173913044 black 
\sparkdot 0.1 0.683391304347826 lightblue 
\sparkdot 0.2 0.606573913043478 lightblue 
\sparkdot 0.3 0.528782608695652 lightblue 
\sparkdot 0.4 0.451669565217391 lightblue 
\sparkdot 0.5 0.372608695652174 lightblue 
\sparkdot 0.6 0.294591304347826 lightblue 
\sparkdot 0.7 0.216121739130435 lightblue 
\sparkdot 0.8 0.139008695652174 lightblue 
\sparkdot 0.9 0.0608173913043478 lightblue 
\sparkdot 1 -0.0188173913043478 blue 
\end{sparkline} & \begin{sparkline}{10}
\spark 0 0.747826086956522 0.1 0.747826086956522 0.2 0.751304347826087 0.3 0.749565217391304 0.4 0.747826086956522 0.5 0.749565217391304 0.6 0.747826086956522 0.7 0.746086956521739 0.8 0.747826086956522 0.9 0.746086956521739 1 0.746086956521739 /
\sparkdot 0 0.74824347826087 red 
\sparkdot 0.1 0.747373913043478 lightblue 
\sparkdot 0.2 0.750904347826087 black 
\sparkdot 0.3 0.749182608695652 lightblue 
\sparkdot 0.4 0.747826086956522 lightblue 
\sparkdot 0.5 0.749217391304348 lightblue 
\sparkdot 0.6 0.747217391304348 lightblue 
\sparkdot 0.7 0.7468 lightblue 
\sparkdot 0.8 0.747495652173913 lightblue 
\sparkdot 0.9 0.746678260869565 lightblue 
\sparkdot 1 0.746521739130435 blue 
\end{sparkline} & \begin{sparkline}{10}
\spark 0 0.511304347826087 0.1 0.511304347826087 0.2 0.511304347826087 0.3 0.511304347826087 0.4 0.509565217391304 0.5 0.509565217391304 0.6 0.507826086956522 0.7 0.511304347826087 0.8 0.507826086956522 0.9 0.506086956521739 1 0.504347826086956 /
\sparkdot 0 0.510521739130435 red 
\sparkdot 0.1 0.510852173913043 lightblue 
\sparkdot 0.2 0.511860869565217 black 
\sparkdot 0.3 0.510869565217391 lightblue 
\sparkdot 0.4 0.50944347826087 lightblue 
\sparkdot 0.5 0.509930434782609 lightblue 
\sparkdot 0.6 0.5084 lightblue 
\sparkdot 0.7 0.510834782608696 lightblue 
\sparkdot 0.8 0.50864347826087 lightblue 
\sparkdot 0.9 0.505478260869565 lightblue 
\sparkdot 1 0.504921739130435 blue 
\end{sparkline} & \begin{sparkline}{10}
\spark 0 0.492173913043478 0.1 0.492173913043478 0.2 0.490434782608696 0.3 0.488695652173913 0.4 0.481739130434783 0.5 0.481739130434783 0.6 0.464347826086957 0.7 0.464347826086957 0.8 0.441739130434783 0.9 0.380869565217391 1 0.267826086956522 /
\sparkdot 0 0.492034782608696 black 
\sparkdot 0.1 0.491634782608696 lightblue 
\sparkdot 0.2 0.489756521739131 lightblue 
\sparkdot 0.3 0.489547826086957 lightblue 
\sparkdot 0.4 0.481165217391304 lightblue 
\sparkdot 0.5 0.481791304347826 lightblue 
\sparkdot 0.6 0.464365217391304 lightblue 
\sparkdot 0.7 0.464695652173913 lightblue 
\sparkdot 0.8 0.442347826086956 lightblue 
\sparkdot 0.9 0.380504347826087 lightblue 
\sparkdot 1 0.268086956521739 blue 
\end{sparkline} & \begin{sparkline}{10}
\spark 0 0.499130434782609 0.1 0.502608695652174 0.2 0.492173913043478 0.3 0.497391304347826 0.4 0.490434782608696 0.5 0.481739130434783 0.6 0.485217391304348 0.7 0.490434782608696 0.8 0.471304347826087 0.9 0.474782608695652 1 0.466086956521739 /
\sparkdot 0 0.49944347826087 red 
\sparkdot 0.1 0.501739130434783 black 
\sparkdot 0.2 0.491826086956522 lightblue 
\sparkdot 0.3 0.496608695652174 lightblue 
\sparkdot 0.4 0.4896 lightblue 
\sparkdot 0.5 0.48104347826087 lightblue 
\sparkdot 0.6 0.485913043478261 lightblue 
\sparkdot 0.7 0.490313043478261 lightblue 
\sparkdot 0.8 0.471408695652174 lightblue 
\sparkdot 0.9 0.474469565217391 lightblue 
\sparkdot 1 0.46575652173913 blue 
\end{sparkline}\\
\hline
German & \begin{sparkline}{10}
\spark 0 0.589565217391304 0.1 0.589565217391304 0.2 0.587826086956522 0.3 0.587826086956522 0.4 0.584347826086956 0.5 0.586086956521739 0.6 0.584347826086956 0.7 0.582608695652174 0.8 0.579130434782609 0.9 0.579130434782609 1 0.577391304347826 /
\sparkdot 0 0.588695652173913 black 
\sparkdot 0.1 0.588695652173913 black 
\sparkdot 0.2 0.588173913043478 lightblue 
\sparkdot 0.3 0.58695652173913 lightblue 
\sparkdot 0.4 0.58504347826087 lightblue 
\sparkdot 0.5 0.585217391304348 lightblue 
\sparkdot 0.6 0.584 lightblue 
\sparkdot 0.7 0.581739130434783 lightblue 
\sparkdot 0.8 0.579652173913043 lightblue 
\sparkdot 0.9 0.579130434782609 lightblue 
\sparkdot 1 0.57704347826087 blue 
\end{sparkline} & \begin{sparkline}{10}
\spark 0 0.546086956521739 0.1 0.516521739130435 0.2 0.48 0.3 0.443478260869565 0.4 0.408695652173913 0.5 0.373913043478261 0.6 0.340869565217391 0.7 0.304347826086956 0.8 0.269565217391304 0.9 0.234782608695652 1 0.2 /
\sparkdot 0 0.546660869565217 black 
\sparkdot 0.1 0.51664347826087 lightblue 
\sparkdot 0.2 0.480869565217391 lightblue 
\sparkdot 0.3 0.444295652173913 lightblue 
\sparkdot 0.4 0.408991304347826 lightblue 
\sparkdot 0.5 0.374608695652174 lightblue 
\sparkdot 0.6 0.340869565217391 lightblue 
\sparkdot 0.7 0.304695652173913 lightblue 
\sparkdot 0.8 0.269043478260869 lightblue 
\sparkdot 0.9 0.23455652173913 lightblue 
\sparkdot 1 0.199773913043478 blue 
\end{sparkline} & \begin{sparkline}{10}
\spark 0 0.579130434782609 0.1 0.579130434782609 0.2 0.582608695652174 0.3 0.570434782608696 0.4 0.577391304347826 0.5 0.579130434782609 0.6 0.580869565217391 0.7 0.565217391304348 0.8 0.568695652173913 0.9 0.573913043478261 1 0.580869565217391 /
\sparkdot 0 0.579478260869565 red 
\sparkdot 0.1 0.579478260869565 lightblue 
\sparkdot 0.2 0.581739130434783 black 
\sparkdot 0.3 0.570434782608696 lightblue 
\sparkdot 0.4 0.576521739130435 lightblue 
\sparkdot 0.5 0.579304347826087 lightblue 
\sparkdot 0.6 0.581565217391304 lightblue 
\sparkdot 0.7 0.565913043478261 lightblue 
\sparkdot 0.8 0.568347826086957 lightblue 
\sparkdot 0.9 0.574260869565217 lightblue 
\sparkdot 1 0.581217391304348 blue 
\end{sparkline} & \begin{sparkline}{10}
\spark 0 0.55304347826087 0.1 0.55304347826087 0.2 0.554782608695652 0.3 0.556521739130435 0.4 0.558260869565217 0.5 0.56 0.6 0.556521739130435 0.7 0.558260869565217 0.8 0.56 0.9 0.556521739130435 1 0.55304347826087 /
\sparkdot 0 0.553652173913044 red 
\sparkdot 0.1 0.553652173913044 lightblue 
\sparkdot 0.2 0.554626086956522 lightblue 
\sparkdot 0.3 0.555808695652174 lightblue 
\sparkdot 0.4 0.558347826086956 lightblue 
\sparkdot 0.5 0.560486956521739 lightblue 
\sparkdot 0.6 0.556678260869565 lightblue 
\sparkdot 0.7 0.559026086956522 lightblue 
\sparkdot 0.8 0.560643478260869 black 
\sparkdot 0.9 0.555704347826087 lightblue 
\sparkdot 1 0.55264347826087 blue 
\end{sparkline} & \begin{sparkline}{10}
\spark 0 0.518260869565217 0.1 0.513043478260869 0.2 0.493913043478261 0.3 0.493913043478261 0.4 0.495652173913044 0.5 0.466086956521739 0.6 0.455652173913044 0.7 0.450434782608696 0.8 0.438260869565217 0.9 0.401739130434783 1 0.35304347826087 /
\sparkdot 0 0.517686956521739 black 
\sparkdot 0.1 0.512817391304348 lightblue 
\sparkdot 0.2 0.4932 lightblue 
\sparkdot 0.3 0.493965217391304 lightblue 
\sparkdot 0.4 0.495686956521739 lightblue 
\sparkdot 0.5 0.466626086956522 lightblue 
\sparkdot 0.6 0.456104347826087 lightblue 
\sparkdot 0.7 0.449686956521739 lightblue 
\sparkdot 0.8 0.438382608695652 lightblue 
\sparkdot 0.9 0.401617391304348 lightblue 
\sparkdot 1 0.352313043478261 blue 
\end{sparkline} & \begin{sparkline}{10}
\spark 0 0.516521739130435 0.1 0.52 0.2 0.52695652173913 0.3 0.518260869565217 0.4 0.504347826086956 0.5 0.495652173913044 0.6 0.476521739130435 0.7 0.488695652173913 0.8 0.47304347826087 0.9 0.455652173913044 1 0.497391304347826 /
\sparkdot 0 0.516017391304348 red 
\sparkdot 0.1 0.520121739130435 lightblue 
\sparkdot 0.2 0.527182608695652 black 
\sparkdot 0.3 0.518921739130435 lightblue 
\sparkdot 0.4 0.504382608695652 lightblue 
\sparkdot 0.5 0.49504347826087 lightblue 
\sparkdot 0.6 0.4768 lightblue 
\sparkdot 0.7 0.4884 lightblue 
\sparkdot 0.8 0.473095652173913 lightblue 
\sparkdot 0.9 0.455704347826087 lightblue 
\sparkdot 1 0.49735652173913 blue 
\end{sparkline}\\
\hline
LoanDefault & \begin{sparkline}{10}
\spark 0 0.709565217391304 0.1 0.709565217391304 0.2 0.707826086956522 0.3 0.706086956521739 0.4 0.704347826086957 0.5 0.704347826086957 0.6 0.702608695652174 0.7 0.700869565217391 0.8 0.699130434782609 0.9 0.697391304347826 1 0.697391304347826 /
\sparkdot 0 0.710052173913044 black 
\sparkdot 0.1 0.708921739130435 lightblue 
\sparkdot 0.2 0.707565217391304 lightblue 
\sparkdot 0.3 0.70615652173913 lightblue 
\sparkdot 0.4 0.704939130434783 lightblue 
\sparkdot 0.5 0.703582608695652 lightblue 
\sparkdot 0.6 0.702278260869565 lightblue 
\sparkdot 0.7 0.700904347826087 lightblue 
\sparkdot 0.8 0.699547826086956 lightblue 
\sparkdot 0.9 0.698104347826087 lightblue 
\sparkdot 1 0.6968 blue 
\end{sparkline} & \begin{sparkline}{10}
\spark 0 0.659130434782609 0.1 0.601739130434783 0.2 0.546086956521739 0.3 0.488695652173913 0.4 0.43304347826087 0.5 0.375652173913043 0.6 0.32 0.7 0.262608695652174 0.8 0.20695652173913 0.9 0.149565217391304 1 0.0921739130434783 /
\sparkdot 0 0.658434782608696 black 
\sparkdot 0.1 0.601930434782609 lightblue 
\sparkdot 0.2 0.545826086956522 lightblue 
\sparkdot 0.3 0.489460869565217 lightblue 
\sparkdot 0.4 0.432747826086957 lightblue 
\sparkdot 0.5 0.376504347826087 lightblue 
\sparkdot 0.6 0.320086956521739 lightblue 
\sparkdot 0.7 0.263165217391304 lightblue 
\sparkdot 0.8 0.206382608695652 lightblue 
\sparkdot 0.9 0.1492 lightblue 
\sparkdot 1 0.0918608695652174 blue 
\end{sparkline} & \begin{sparkline}{10}
\spark 0 0.706086956521739 0.1 0.706086956521739 0.2 0.706086956521739 0.3 0.707826086956522 0.4 0.706086956521739 0.5 0.706086956521739 0.6 0.707826086956522 0.7 0.706086956521739 0.8 0.706086956521739 0.9 0.706086956521739 1 0.706086956521739 /
\sparkdot 0 0.706365217391304 red 
\sparkdot 0.1 0.706904347826087 lightblue 
\sparkdot 0.2 0.706226086956522 lightblue 
\sparkdot 0.3 0.706991304347826 lightblue 
\sparkdot 0.4 0.706486956521739 lightblue 
\sparkdot 0.5 0.705721739130435 lightblue 
\sparkdot 0.6 0.708 black 
\sparkdot 0.7 0.705756521739131 lightblue 
\sparkdot 0.8 0.706904347826087 lightblue 
\sparkdot 0.9 0.706521739130435 lightblue 
\sparkdot 1 0.706173913043478 blue 
\end{sparkline} & \begin{sparkline}{10}
\spark 0 0.798260869565217 0.1 0.798260869565217 0.2 0.798260869565217 0.3 0.798260869565217 0.4 0.796521739130435 0.5 0.796521739130435 0.6 0.796521739130435 0.7 0.796521739130435 0.8 0.796521739130435 0.9 0.796521739130435 1 0.796521739130435 /
\sparkdot 0 0.797617391304348 red 
\sparkdot 0.1 0.79775652173913 lightblue 
\sparkdot 0.2 0.797617391304348 lightblue 
\sparkdot 0.3 0.797808695652174 black 
\sparkdot 0.4 0.797095652173913 lightblue 
\sparkdot 0.5 0.797130434782609 lightblue 
\sparkdot 0.6 0.796904347826087 lightblue 
\sparkdot 0.7 0.796904347826087 lightblue 
\sparkdot 0.8 0.796765217391304 lightblue 
\sparkdot 0.9 0.796973913043478 lightblue 
\sparkdot 1 0.796608695652174 blue 
\end{sparkline} & \begin{sparkline}{10}
\spark 0 0.801739130434783 0.1 0.805217391304348 0.2 0.803478260869565 0.3 0.8 0.4 0.8 0.5 0.796521739130435 0.6 0.791304347826087 0.7 0.79304347826087 0.8 0.786086956521739 0.9 0.777391304347826 1 0.419130434782609 /
\sparkdot 0 0.80144347826087 red 
\sparkdot 0.1 0.804539130434783 black 
\sparkdot 0.2 0.802747826086957 lightblue 
\sparkdot 0.3 0.800434782608696 lightblue 
\sparkdot 0.4 0.799634782608696 lightblue 
\sparkdot 0.5 0.796469565217391 lightblue 
\sparkdot 0.6 0.792139130434783 lightblue 
\sparkdot 0.7 0.792539130434783 lightblue 
\sparkdot 0.8 0.785913043478261 lightblue 
\sparkdot 0.9 0.777982608695652 lightblue 
\sparkdot 1 0.418713043478261 blue 
\end{sparkline} & \begin{sparkline}{10}
\spark 0 0.803478260869565 0.1 0.79304347826087 0.2 0.787826086956522 0.3 0.796521739130435 0.4 0.786086956521739 0.5 0.787826086956522 0.6 0.779130434782609 0.7 0.780869565217391 0.8 0.775652173913043 0.9 0.779130434782609 1 0.775652173913043 /
\sparkdot 0 0.802608695652174 black 
\sparkdot 0.1 0.792921739130435 lightblue 
\sparkdot 0.2 0.787791304347826 lightblue 
\sparkdot 0.3 0.797113043478261 lightblue 
\sparkdot 0.4 0.785478260869565 lightblue 
\sparkdot 0.5 0.788504347826087 lightblue 
\sparkdot 0.6 0.779373913043478 lightblue 
\sparkdot 0.7 0.780191304347826 lightblue 
\sparkdot 0.8 0.775634782608696 lightblue 
\sparkdot 0.9 0.779704347826087 lightblue 
\sparkdot 1 0.775147826086956 blue 
\end{sparkline}\\
\hline
MedicalExpenditure & \begin{sparkline}{10}
\spark 0 0.779130434782609 0.1 0.772173913043478 0.2 0.765217391304348 0.3 0.756521739130435 0.4 0.747826086956522 0.5 0.740869565217391 0.6 0.733913043478261 0.7 0.725217391304348 0.8 0.718260869565217 0.9 0.709565217391304 1 0.702608695652174 /
\sparkdot 0 0.7796 black 
\sparkdot 0.1 0.77224347826087 lightblue 
\sparkdot 0.2 0.764382608695652 lightblue 
\sparkdot 0.3 0.756608695652174 lightblue 
\sparkdot 0.4 0.748626086956522 lightblue 
\sparkdot 0.5 0.741182608695652 lightblue 
\sparkdot 0.6 0.733321739130435 lightblue 
\sparkdot 0.7 0.725704347826087 lightblue 
\sparkdot 0.8 0.71784347826087 lightblue 
\sparkdot 0.9 0.71 lightblue 
\sparkdot 1 0.702069565217391 blue 
\end{sparkline} & \begin{sparkline}{10}
\spark 0 0.76695652173913 0.1 0.690434782608696 0.2 0.613913043478261 0.3 0.537391304347826 0.4 0.460869565217391 0.5 0.386086956521739 0.6 0.309565217391304 0.7 0.23304347826087 0.8 0.158260869565217 0.9 0.0817391304347826 1 0.00521739130434783 /
\sparkdot 0 0.766139130434783 black 
\sparkdot 0.1 0.690086956521739 lightblue 
\sparkdot 0.2 0.613652173913043 lightblue 
\sparkdot 0.3 0.537513043478261 lightblue 
\sparkdot 0.4 0.461217391304348 lightblue 
\sparkdot 0.5 0.385582608695652 lightblue 
\sparkdot 0.6 0.309895652173913 lightblue 
\sparkdot 0.7 0.233373913043478 lightblue 
\sparkdot 0.8 0.157478260869565 lightblue 
\sparkdot 0.9 0.0815652173913043 lightblue 
\sparkdot 1 0.00493913043478253 blue 
\end{sparkline} & \begin{sparkline}{10}
\spark 0 0.777391304347826 0.1 0.777391304347826 0.2 0.779130434782609 0.3 0.777391304347826 0.4 0.777391304347826 0.5 0.773913043478261 0.6 0.775652173913043 0.7 0.775652173913043 0.8 0.775652173913043 0.9 0.773913043478261 1 0.773913043478261 /
\sparkdot 0 0.776556521739131 red 
\sparkdot 0.1 0.777095652173913 lightblue 
\sparkdot 0.2 0.778330434782609 black 
\sparkdot 0.3 0.776765217391304 lightblue 
\sparkdot 0.4 0.777895652173913 lightblue 
\sparkdot 0.5 0.774017391304348 lightblue 
\sparkdot 0.6 0.775460869565217 lightblue 
\sparkdot 0.7 0.7748 lightblue 
\sparkdot 0.8 0.774939130434783 lightblue 
\sparkdot 0.9 0.773634782608696 lightblue 
\sparkdot 1 0.773756521739131 blue 
\end{sparkline} & \begin{sparkline}{10}
\spark 0 0.111304347826087 0.1 0.11304347826087 0.2 0.11304347826087 0.3 0.11304347826087 0.4 0.114782608695652 0.5 0.11304347826087 0.6 0.114782608695652 0.7 0.11304347826087 0.8 0.114782608695652 0.9 0.114782608695652 1 0.116521739130435 /
\sparkdot 0 0.111913043478261 red 
\sparkdot 0.1 0.112295652173913 lightblue 
\sparkdot 0.2 0.1132 lightblue 
\sparkdot 0.3 0.113339130434783 lightblue 
\sparkdot 0.4 0.114713043478261 lightblue 
\sparkdot 0.5 0.113113043478261 lightblue 
\sparkdot 0.6 0.113965217391304 lightblue 
\sparkdot 0.7 0.113373913043478 lightblue 
\sparkdot 0.8 0.11415652173913 lightblue 
\sparkdot 0.9 0.115008695652174 lightblue 
\sparkdot 1 0.115686956521739 black 
\end{sparkline} & \begin{sparkline}{10}
\spark 0 0.11304347826087 0.1 0.11304347826087 0.2 0.116521739130435 0.3 0.12 0.4 0.121739130434783 0.5 0.12 0.6 0.118260869565217 0.7 0.118260869565217 0.8 0.123478260869565 0.9 0.12 1 0.106086956521739 /
\sparkdot 0 0.112278260869565 red 
\sparkdot 0.1 0.11384347826087 lightblue 
\sparkdot 0.2 0.116608695652174 lightblue 
\sparkdot 0.3 0.119252173913043 lightblue 
\sparkdot 0.4 0.121478260869565 lightblue 
\sparkdot 0.5 0.119391304347826 lightblue 
\sparkdot 0.6 0.118521739130435 lightblue 
\sparkdot 0.7 0.117495652173913 lightblue 
\sparkdot 0.8 0.12415652173913 black 
\sparkdot 0.9 0.120226086956522 lightblue 
\sparkdot 1 0.106608695652174 blue 
\end{sparkline} & \begin{sparkline}{10}
\spark 0 0.106086956521739 0.1 0.100869565217391 0.2 0.0852173913043478 0.3 0.0817391304347826 0.4 0.0678260869565217 0.5 0.0626086956521739 0.6 0.0539130434782608 0.7 0.0452173913043478 0.8 0.0399999999999999 0.9 0.0399999999999999 1 0.0226086956521738 /
\sparkdot 0 0.106886956521739 black 
\sparkdot 0.1 0.101078260869565 lightblue 
\sparkdot 0.2 0.0853217391304348 lightblue 
\sparkdot 0.3 0.0819304347826087 lightblue 
\sparkdot 0.4 0.0685565217391304 lightblue 
\sparkdot 0.5 0.0622434782608696 lightblue 
\sparkdot 0.6 0.0543652173913043 lightblue 
\sparkdot 0.7 0.044695652173913 lightblue 
\sparkdot 0.8 0.0400521739130435 lightblue 
\sparkdot 0.9 0.0393043478260869 lightblue 
\sparkdot 1 0.0224869565217391 blue 
\end{sparkline}\\
\hline
Portuguese & \begin{sparkline}{10}
\spark 0 0.883478260869565 0.1 0.883478260869565 0.2 0.883478260869565 0.3 0.883478260869565 0.4 0.883478260869565 0.5 0.883478260869565 0.6 0.883478260869565 0.7 0.883478260869565 0.8 0.883478260869565 0.9 0.883478260869565 1 0.883478260869565 /
\sparkdot 0 0.882765217391304 black 
\sparkdot 0.1 0.882765217391304 black 
\sparkdot 0.2 0.882765217391304 black 
\sparkdot 0.3 0.882765217391304 black 
\sparkdot 0.4 0.882765217391304 black 
\sparkdot 0.5 0.882765217391304 black 
\sparkdot 0.6 0.882765217391304 black 
\sparkdot 0.7 0.882765217391304 black 
\sparkdot 0.8 0.882765217391304 black 
\sparkdot 0.9 0.882765217391304 black 
\sparkdot 1 0.882765217391304 black 
\end{sparkline} & \begin{sparkline}{10}
\spark 0 0.881739130434783 0.1 0.881739130434783 0.2 0.881739130434783 0.3 0.881739130434783 0.4 0.881739130434783 0.5 0.881739130434783 0.6 0.881739130434783 0.7 0.881739130434783 0.8 0.881739130434783 0.9 0.881739130434783 1 0.881739130434783 /
\sparkdot 0 0.881826086956522 black 
\sparkdot 0.1 0.881826086956522 black 
\sparkdot 0.2 0.881826086956522 black 
\sparkdot 0.3 0.881826086956522 black 
\sparkdot 0.4 0.881826086956522 black 
\sparkdot 0.5 0.881773913043478 lightblue 
\sparkdot 0.6 0.881739130434783 lightblue 
\sparkdot 0.7 0.881704347826087 lightblue 
\sparkdot 0.8 0.881686956521739 lightblue 
\sparkdot 0.9 0.881669565217391 lightblue 
\sparkdot 1 0.881617391304348 blue 
\end{sparkline} & \begin{sparkline}{10}
\spark 0 0.88 0.1 0.88 0.2 0.88 0.3 0.881739130434783 0.4 0.88 0.5 0.881739130434783 0.6 0.88 0.7 0.88 0.8 0.88 0.9 0.881739130434783 1 0.88 /
\sparkdot 0 0.880069565217391 red 
\sparkdot 0.1 0.880695652173913 lightblue 
\sparkdot 0.2 0.880539130434783 lightblue 
\sparkdot 0.3 0.880973913043478 lightblue 
\sparkdot 0.4 0.880295652173913 lightblue 
\sparkdot 0.5 0.880939130434783 lightblue 
\sparkdot 0.6 0.880208695652174 lightblue 
\sparkdot 0.7 0.880382608695652 lightblue 
\sparkdot 0.8 0.880486956521739 lightblue 
\sparkdot 0.9 0.881060869565217 black 
\sparkdot 1 0.879826086956522 blue 
\end{sparkline} & \begin{sparkline}{10}
\spark 0 0.986086956521739 0.1 0.986086956521739 0.2 0.986086956521739 0.3 0.986086956521739 0.4 0.986086956521739 0.5 0.986086956521739 0.6 0.986086956521739 0.7 0.986086956521739 0.8 0.986086956521739 0.9 0.986086956521739 1 0.986086956521739 /
\sparkdot 0 0.986139130434783 black 
\sparkdot 0.1 0.986139130434783 black 
\sparkdot 0.2 0.986139130434783 black 
\sparkdot 0.3 0.986139130434783 black 
\sparkdot 0.4 0.986139130434783 black 
\sparkdot 0.5 0.986139130434783 black 
\sparkdot 0.6 0.986139130434783 black 
\sparkdot 0.7 0.986139130434783 black 
\sparkdot 0.8 0.986139130434783 black 
\sparkdot 0.9 0.986139130434783 black 
\sparkdot 1 0.986139130434783 black 
\end{sparkline} & \begin{sparkline}{10}
\spark 0 0.998260869565217 0.1 0.998260869565217 0.2 0.998260869565217 0.3 0.998260869565217 0.4 0.998260869565217 0.5 0.998260869565217 0.6 0.998260869565217 0.7 0.998260869565217 0.8 0.998260869565217 0.9 0.998260869565217 1 0.998260869565217 /
\sparkdot 0 0.997965217391304 red 
\sparkdot 0.1 0.997965217391304 lightblue 
\sparkdot 0.2 0.997965217391304 lightblue 
\sparkdot 0.3 0.997965217391304 lightblue 
\sparkdot 0.4 0.997965217391304 lightblue 
\sparkdot 0.5 0.997686956521739 lightblue 
\sparkdot 0.6 0.998121739130435 lightblue 
\sparkdot 0.7 0.998121739130435 lightblue 
\sparkdot 0.8 0.99824347826087 black 
\sparkdot 0.9 0.998208695652174 lightblue 
\sparkdot 1 0.998104347826087 blue 
\end{sparkline} & \begin{sparkline}{10}
\spark 0 0.987826086956522 0.1 0.987826086956522 0.2 0.989565217391304 0.3 0.99304347826087 0.4 0.996521739130435 0.5 1 0.6 1.00173913043478 0.7 1.00347826086957 0.8 1.00695652173913 0.9 1.00869565217391 1 1.0104347826087 /
\sparkdot 0 0.987547826086957 red 
\sparkdot 0.1 0.987947826086957 lightblue 
\sparkdot 0.2 0.990313043478261 lightblue 
\sparkdot 0.3 0.9936 lightblue 
\sparkdot 0.4 0.996469565217391 lightblue 
\sparkdot 0.5 0.999391304347826 lightblue 
\sparkdot 0.6 1.00147826086957 lightblue 
\sparkdot 0.7 1.00302608695652 lightblue 
\sparkdot 0.8 1.00610434782609 lightblue 
\sparkdot 0.9 1.00900869565217 lightblue 
\sparkdot 1 1.0104347826087 black 
\end{sparkline}\\
\hline
StudentPerformance & \begin{sparkline}{10}
\spark 0 0.389565217391304 0.1 0.389565217391304 0.2 0.389565217391304 0.3 0.389565217391304 0.4 0.389565217391304 0.5 0.391304347826087 0.6 0.389565217391304 0.7 0.389565217391304 0.8 0.389565217391304 0.9 0.389565217391304 1 0.387826086956522 /
\sparkdot 0 0.389234782608696 red 
\sparkdot 0.1 0.389234782608696 lightblue 
\sparkdot 0.2 0.389234782608696 lightblue 
\sparkdot 0.3 0.389234782608696 lightblue 
\sparkdot 0.4 0.390104347826087 lightblue 
\sparkdot 0.5 0.390973913043478 black 
\sparkdot 0.6 0.389652173913043 lightblue 
\sparkdot 0.7 0.389217391304348 lightblue 
\sparkdot 0.8 0.389634782608696 lightblue 
\sparkdot 0.9 0.388747826086957 lightblue 
\sparkdot 1 0.388278260869565 blue 
\end{sparkline} & \begin{sparkline}{10}
\spark 0 0.389565217391304 0.1 0.389565217391304 0.2 0.379130434782609 0.3 0.365217391304348 0.4 0.35304347826087 0.5 0.342608695652174 0.6 0.330434782608696 0.7 0.318260869565217 0.8 0.300869565217391 0.9 0.288695652173913 1 0.28 /
\sparkdot 0 0.390139130434783 black 
\sparkdot 0.1 0.38984347826087 lightblue 
\sparkdot 0.2 0.379321739130435 lightblue 
\sparkdot 0.3 0.365513043478261 lightblue 
\sparkdot 0.4 0.352191304347826 lightblue 
\sparkdot 0.5 0.342608695652174 lightblue 
\sparkdot 0.6 0.330226086956522 lightblue 
\sparkdot 0.7 0.3176 lightblue 
\sparkdot 0.8 0.300573913043478 lightblue 
\sparkdot 0.9 0.288852173913043 lightblue 
\sparkdot 1 0.280486956521739 blue 
\end{sparkline} & \begin{sparkline}{10}
\spark 0 0.349565217391304 0.1 0.346086956521739 0.2 0.332173913043478 0.3 0.330434782608696 0.4 0.328695652173913 0.5 0.323478260869565 0.6 0.349565217391304 0.7 0.351304347826087 0.8 0.354782608695652 0.9 0.344347826086956 1 0.354782608695652 /
\sparkdot 0 0.348852173913043 red 
\sparkdot 0.1 0.3456 lightblue 
\sparkdot 0.2 0.333026086956522 lightblue 
\sparkdot 0.3 0.329878260869565 lightblue 
\sparkdot 0.4 0.328695652173913 lightblue 
\sparkdot 0.5 0.323878260869565 lightblue 
\sparkdot 0.6 0.349547826086957 lightblue 
\sparkdot 0.7 0.351234782608696 lightblue 
\sparkdot 0.8 0.355426086956522 black 
\sparkdot 0.9 0.345165217391304 lightblue 
\sparkdot 1 0.354052173913043 blue 
\end{sparkline} & \begin{sparkline}{10}
\spark 0 0.530434782608696 0.1 0.530434782608696 0.2 0.530434782608696 0.3 0.530434782608696 0.4 0.530434782608696 0.5 0.52695652173913 0.6 0.528695652173913 0.7 0.533913043478261 0.8 0.533913043478261 0.9 0.535652173913043 1 0.528695652173913 /
\sparkdot 0 0.5312 red 
\sparkdot 0.1 0.5312 lightblue 
\sparkdot 0.2 0.5312 lightblue 
\sparkdot 0.3 0.5312 lightblue 
\sparkdot 0.4 0.530852173913043 lightblue 
\sparkdot 0.5 0.527547826086956 lightblue 
\sparkdot 0.6 0.529286956521739 lightblue 
\sparkdot 0.7 0.533147826086956 lightblue 
\sparkdot 0.8 0.533878260869565 lightblue 
\sparkdot 0.9 0.535478260869565 black 
\sparkdot 1 0.52784347826087 blue 
\end{sparkline} & \begin{sparkline}{10}
\spark 0 0.48 0.1 0.47304347826087 0.2 0.47304347826087 0.3 0.471304347826087 0.4 0.450434782608696 0.5 0.438260869565217 0.6 0.401739130434783 0.7 0.368695652173913 0.8 0.349565217391304 0.9 0.323478260869565 1 0.288695652173913 /
\sparkdot 0 0.479391304347826 black 
\sparkdot 0.1 0.472643478260869 lightblue 
\sparkdot 0.2 0.473147826086956 lightblue 
\sparkdot 0.3 0.470886956521739 lightblue 
\sparkdot 0.4 0.450052173913044 lightblue 
\sparkdot 0.5 0.438034782608696 lightblue 
\sparkdot 0.6 0.400991304347826 lightblue 
\sparkdot 0.7 0.3688 lightblue 
\sparkdot 0.8 0.350086956521739 lightblue 
\sparkdot 0.9 0.3228 lightblue 
\sparkdot 1 0.289304347826087 blue 
\end{sparkline} & \begin{sparkline}{10}
\spark 0 0.544347826086956 0.1 0.624347826086956 0.2 0.580869565217391 0.3 0.579130434782609 0.4 0.60695652173913 0.5 0.605217391304348 0.6 0.558260869565217 0.7 0.579130434782609 0.8 0.608695652173913 0.9 0.572173913043478 1 0.55304347826087 /
\sparkdot 0 0.543878260869565 red 
\sparkdot 0.1 0.623913043478261 black 
\sparkdot 0.2 0.581426086956522 lightblue 
\sparkdot 0.3 0.579617391304348 lightblue 
\sparkdot 0.4 0.606330434782609 lightblue 
\sparkdot 0.5 0.605860869565217 lightblue 
\sparkdot 0.6 0.558330434782609 lightblue 
\sparkdot 0.7 0.579095652173913 lightblue 
\sparkdot 0.8 0.609234782608696 lightblue 
\sparkdot 0.9 0.57224347826087 lightblue 
\sparkdot 1 0.553060869565217 blue 
\end{sparkline}\\
\hline
\end{tabular}

\FloatBarrier

\bibliography{bib}
\bibliographystyle{icml2021}